\documentclass{article} 
\usepackage[nonatbib,final]{arxiv_nips_2016}
\usepackage[svgnames]{xcolor}
\usepackage{url}
\usepackage{amsmath,amssymb}
\usepackage{algorithm}
\usepackage{algpseudocode}
\usepackage{graphicx}
\usepackage{wrapfig}
\usepackage{array}
\usepackage{multicol}
\usepackage{caption}  
\usepackage{tikz}
\usepackage{subcaption} 
\usepackage{mathtools}
\usepackage{hyperref}

\usetikzlibrary{shapes,arrows}
\definecolor{Gray}{gray}{0.9}

\pgfdeclarelayer{edgelayer}
\pgfdeclarelayer{nodelayer}
\pgfsetlayers{edgelayer,nodelayer,main}

\tikzstyle{none}=[inner sep=2pt]

\tikzstyle{vertex_default}=[circle,fill=White,draw=Black,text width=0.1cm]
\tikzstyle{tree_edge}=[thick,draw=Black]
\tikzstyle{add_edge}=[thick,draw=Red]
\tikzstyle{local_edge}=[thick,draw=Blue]

\title{Efficient Pose and Cell Segmentation using Column Generation}
\author{
Shaofei Wang\\
Beijing China \\
\texttt{sfwang0928@gmail.com} 
\And
Chong Zhang\\
SIMBioSys Group, DTIC, Universitat Pompeu Fabra \\
Barcelona Spain \\
\texttt{chong.zhang@upf.edu} 
\And
Miguel A. Gonzalez-Ballester\\
SIMBioSys Group, DTIC, Universitat Pompeu Fabra \& ICREA,\\
Barcelona Spain \\
\texttt{ma.gonzalez@upf.edu} 
\And
Julian Yarkony\\
Experian Data Lab \\
San Diego CA \\
\texttt{julian.e.yarkony@gmail.com} 
}

\begin{document}

\maketitle
\begin{abstract}
We study the problems of multi-person pose segmentation in natural images and instance segmentation in biological images with crowded cells. We formulate these distinct tasks as integer programs where variables correspond to poses/cells. To optimize, we propose a generic relaxation scheme for solving these combinatorial problems using a column generation formulation where the program for generating a column is solved
via exact optimization of very small scale integer programs.  This results in efficient exploration of the spaces of poses and cells.  
\end{abstract}
\section{Introduction}
In this paper we consider two distinct problems: multi-person pose segmentation in natural images and instance segmentation in biological images (bioimages).  Multi-person pose segmentation is the problem of detecting people and their corresponding body parts in natural images.  Each pose provides a description of the positions of the body parts of a given person.  Instance segmentation in bioimages is the problem of detecting and segmenting individuals in crowded clustered cells. In both domains poses and cells are in close proximity and often occlude each other.  

We formulate the problem  of multi-person pose segmentation from the perspective of selecting a subset of high-quality poses subject to the constraint that no pair of selected poses is associated with a common  body part detection.  Similarly we formulate the problem of cell instance segmentation from the perspective of grouping super-pixels into cells subject to the constraint that no two cells share a common super-pixel.   

We present a relaxation of these combinatorial problem that uses a column generation formulation for inference where columns are generated  via solving small scale, tractable integer linear programs (ILP). Our work builds on initial ideas of~\cite{yarkoNips2016}.  In this paper the authors formulate the multiple object tracking problem as a maximum weight set packing problem \cite{karp}, where track costs are derived from high-order Markov models.  Tracks are generated in large quantities using dynamic programming. The corresponding LP relaxation is generally tight but the authors further tighten it using third order odd set inequalities \cite{heismann2014generalization}.  Our work differs from \cite{yarkoNips2016} with regards to the problems selected and how column generation is formulated/applied.  Our model is unique in that it allows for the expression of a combinatorial number of poses using a global-local structure.

Our application on multi-person pose segmentation is based on the work of \cite{deepcut1,deepcut2} which casts multi-person pose segmentation as a massive ILP.  They tackle inference using a state of the art ILP solver, assisted with greedy multi-stage optimization.  
We rely on column generation \cite{cuttingstock,barnprice} methods tailored specifically to multi-person pose segmentation. 
Our inference is notably efficient since the dynamic programming inference inspired by the deformable part model of \cite{deva1,deva2,deva3} can be applied to generate columns.  

Our application on cell instance segmentation is related to \cite{ZhangSchwingICCV2015}. In \cite{ZhangSchwingICCV2015} the authors use depth to transform instance segmentation into a labeling problem and thus break the difficult symmetries found when studying the problem.  They frame optimization as an ILP and attack it using the greedy methods of network flows \cite{boykov2001fast} notably Quadratic Pseudo-Boolean Optimization \cite{rother2007optimizing,boros2002pseudo}.  To solve the optimization problem, \cite{ZhangSchwingICCV2015} requires specific knowledge about the number of labels present in the image, while our work uses an ILP framework that does not require such knowledge.

\cite{Zhang14a,Zhang14b} attacks the problem of  segmenting a large number of objects of similar type in crowded images, where cells touch or overlap with each other. They start with a set of over-segmented super-pixels, which are then clustered into ``perceptually meaningful" regions.  Our work is distinct from \cite{Zhang14b} primarily from the perspective of optimization.  The authors of \cite{Zhang14b} rely heavily on the planarity of their problem structure to permit efficient inference.  We however are not bound by such a restriction and instead rely on the structure of cells being small and compact.  

Our paper is organized as follows.  In Section \ref{probForm} we frame multi-person pose segmentation as optimization.  In Section \ref{solepose} we provide an alternative formulation of the optimization problem that is amenable to efficient inference via column generation.  Next in Section \ref{violposes} we study the generation of variables for use in column generation.  Then in Section \ref{Cells} we apply the material in Section \ref{solepose} to the problem of instance segmentation of cells in dense bioimages.  In Section \ref{lbUb} we discuss the production of anytime upper and lower bounds on the optimal integer solution.  In Section \ref{poseCellExp} we demonstrate state of the art results on multi-person pose segmentation and cell instance segmentation.  Finally we conclude in Section \ref{concsectiondone}.
\section{Modeling Multi-Person Pose Segmentation}
\label{probForm}
We now consider our approach for multi-person pose segmentation.  Given an image we first compute a set of body part detections/key-points using the body part detector of \cite{deepcut1}. Each key-point is associated with exactly one body part. We consider the following fourteen body parts:  head, neck and the left/right of the following; ankle, knee, hip, wrist, elbow, shoulder.  A pose describes a person in an image in terms of the location of their body parts and is defined by a subset of the key-points. 

Given the key-points and the part associated to each key-point, we aim to describe poses of multiple people in the image using these key-points. We formulate this problem as an ILP, as will be outlined in the following sections.  
\subsection{Global-Local Structure}
\label{globloc}
Our model for multi-person pose segmentation is associated with a structure which we refer to as global-local structure.  In global-local structure, key-points are labeled as global or local during optimization.  Parts are labeled as global or local as specified by our model before optimization.
\begin{itemize}
\item \textbf{Global Parts:  }
Global parts are a set of uncommonly occluded parts such as the neck.  A part is occluded in a particular pose if no key-point in that pose is associated with that part.   Our model assumes that the occlusion of all global parts in a given pose is strong evidence for a proposed pose being false. Our model thus requires that every pose is associated with at least one non-occluded global part. %
\item \textbf{Local Parts:  }  A local part is simply a part that is not a global part. Local parts are commonly  occluded. 
\item \textbf{Global Key-Points:  }  Global key-points facilitate the modeling of the concept that key-points associated with a common pose should be consistent with regards to distance/angular relationships associated with the parts in a human.  An example of such a relationships is that the head is above the neck.  For every pose, all non-occluded parts in that pose are associated with exactly one global key-point.   
\item \textbf{Local Key-Points:  }  A group of local key-points is a set of key-points that belong the same part of the same person/pose. Key-points that are visually similar, are in close spatial proximity, and correspond to the same part tend to be associated with a common person, while key-points that correspond to the same part but are not visually similar or in close spatial proximity tend to be associated with different people. The use of local-key points can be understood as modeling non-maximum suppression (NMS) during optimization, instead of having NMS as a pre-processing or post-processing step.  
\end{itemize}
\subsection{Problem Formulation}
In this section we formulate multi-person pose segmentation as an ILP.  We denote the set of key-points as $\mathcal{D}$ which we index with  $d$.  We use $\mathcal{R}$ to denote the set of parts which we index by $r$. We use $\hat{\mathcal{R}}\subseteq \mathcal{R}$ to denote the set of  global parts. We describe the mapping of key-points to parts using a  matrix $R \in \{ 0,1 \}^{|\mathcal{D}|  \times |\mathcal{R}|}$ which we index by $d,r$.  We use $R_{dr}=1$ to indicate that key-point $d$ is associated with part $r$.  For short hand we use $R_{d}$ to indicate the part associated with key-point $d$.  

We use three types of binary indicator variables $x \in \{0,1\}^{|\mathcal{D}|}$, $y \in \{0,1\}^{|\mathcal{D}|\times |\mathcal{D}|}$, 
and $z \in \{0,1\}^{|\mathcal{D}|}$.  We set $x_d=1$ to indicate that key-point $d$ is included in a pose and set $x_d=0$ otherwise. We refer to a key-point $d$ such that $x_d=1$ as being active.  We use $y_{d_1d_2}=1$ to indicate that $d_1$ and $d_2$ are associated with a common pose. We use $z_d=1$ to indicate that the key-point $d$ is labeled as global. If $x_d=1$ and $z_d=0$ then key-point $d$ is associated with the label local.  
%
%
%
%
\subsubsection{Associating Cost to Pose Segmentation}
We now associate a cost function to the problem of multi-person pose segmentation.  To facilitate this we introduce  terms $\theta \in \mathbb{R}^{|\mathcal{D}|}$ and $\phi \in \mathbb{R}^{|\mathcal{D}|\times |\mathcal{D}|}$, which we index by $d$ and $d_1d_2$ respectively.  We use $\theta_d$ to denote the cost of assigning $d$ to a pose (making $d$ active).  We use $\phi_{d_1d_2}$ to denote the cost of assigning $d_1$ and $d_2$ to a common pose.  We associate a cost to a solution $x$,$y$,$z$ via  the terms $\theta$,$\phi$ using a cost function with the following three terms.
\begin{itemize}
\item $\sum_{d\in \mathcal{D}} \theta_d x_d$: This term describes the cost incurred for making key-points active. 
\item $\sum_{\substack{d_1,d_2 \in \mathcal{D}\\R_{d_{1}}=R{d_2}}}\phi_{d_1d_2}y_{d_1d_2}$:  This term describes the cost incurred for  associating key-points of a common part to a shared pose.  
\item $\sum_{\substack{d_1,d_2 \in \mathcal{D}\\R_{d_{1}}\neq R{d_2}}}\phi_{d_1d_2}z_{d_1}z_{d_2}y_{d_1d_2}$: This term describes the cost incurred for associating global key-points with common pose.  
\item We combine the second and third terms into a common term as follows:  
$\sum_{d_1,d_2\in \mathcal{D}}\phi_{d_1d_2}y_{d_1d_2}([R_{d_1}=R_{d_2}]+z_{d_1}z_{d_2})$.  Here $[...]$ is used to express the binary indicator function.
\item Given a solution $x,y,z$ the corresponding cost is written below.  
\begin{align}
\label{mycostdef}
\sum_{d\in \mathcal{D}} \theta_d x_d+\sum_{d_1,d_2\in \mathcal{D}}\phi_{d_1d_2}y_{d_1d_2}([R_{d_1}=R_{d_2}]+z_{d_1}z_{d_2})
\end{align}
\end{itemize}
\subsubsection{Consistency of Solution}
Minimizing the cost in Eq \ref{mycostdef} with respect to $x,y,z$ does not result in consistent poses since the variables $x$, $y$ and $z$ are related to each other. To ensure a consistent solution we enforce the following constraints:
\begin{itemize}
\item 
A key-point that is labeled global has to be active.
 \begin{align}
 \nonumber 
z_d \leq x_d \quad \forall d\in\mathcal{D}
\end{align}
\item 
A key-point pair $d_1,d_2$ can not be assigned to a common pose unless both are active. 
 \begin{align}
 \nonumber 
 y_{d_1d_2} \leq x_{d_1} \quad \forall d_1,d_2 \in \mathcal{D}\\
\nonumber  y_{d_1d_2} \leq x_{d_2}\quad \forall d_1,d_2 \in \mathcal{D}
 \end{align}
\item 
If $d_1$,$d_2$ are associated with a common pose and $d_1$,$d_3$ are associated with a common pose then $d_2$,$d_3$ must be associated with a common pose.
 \begin{align}
 \nonumber
y_{d_1d_2}+y_{d_1,d_3}-y_{d_2,d_3}\leq 1\quad \forall d_1,d_2,d_3 \in \mathcal{D}
\end{align}
\item 
No two global key-points associated with a common part can be  associated with a common pose.  
 \begin{align}
\nonumber
y_{d_1d_2}+z_{d_1}z_{d_2}[R_{d_1}=R_{d_2}]\leq 1 \quad \forall d_1,d_2 \in \mathcal{D}
\end{align}
\item 
Every  active local key-point  of a given part must be associated with an active global key-point of that part.  This ensures that every non-occluded part in a given pose is associated with interactions between all other non-occluded parts in that pose. 
\begin{align}
\nonumber
x_{d_1}-z_{d_1} \leq \sum_{\substack{d_2 \in \mathcal{D} \\ R_{d_2}=R_{d_1}}}y_{d_1d_2}z_{d_2} \quad \forall d_1 \in \mathcal{D}
\end{align}
\item 
Every active  key-point is associated with an active key-point of a global part.  This enforces the constraint that every pose is associated with at least one global part.     
\begin{align}
\nonumber
x_{d_1} \leq \sum_{r \in \hat{\mathcal{R}}}(R_{d_1r}+ \sum_{d_2 \in \mathcal{D}}y_{d_1d_2}z_{d_2}R_{d_2r})   \quad \forall d_1 \in \mathcal{D}
\end{align}
\end{itemize}
Combining all the aforementioned constraints as well as the cost function terms we obtain the following ILP for the multi-person pose segmentation problem:
\begin{align}
\label{eq:MainProg}
\min_{\substack{x\in \{ 0,1\}^{|\mathcal{D}|}\\ %
y  \in \{0,1\}^{|\mathcal{D}|\times |\mathcal{D}|} \\
z  \in \{0,1\}^{|\mathcal{D}|}}} &
\sum_{d\in \mathcal{D}}\theta_dx_d +
 \sum_{\substack{d_1\in \mathcal{D}\\d_2\in \mathcal{D}}}\phi_{d_1d_2}y_{d_1d_2}([R_{d_1}=R_{d_2}]+z_{d_1}z_{d_2})\\
\text{s.t.} \quad &\nonumber z_d\leq x_d, \quad & \forall d \in \mathcal{D}\\
&\nonumber  y_{d_1d_2}\leq x_{d_1}, \quad & \forall d_1,d_2 \in \mathcal{D}\\
&\nonumber y_{d_1d_2}\leq x_{d_2}, \quad & \forall d_1,d_2 \in \mathcal{D}\\
&\nonumber y_{d_1d_2}+y_{d_1,d_3}-y_{d_2,d_3}\leq 1, \quad & \forall d_1,d_2,d_3 \in \mathcal{D}\\
&\nonumber y_{d_1d_2}+z_{d_1}z_{d_2}[R_{d_1}=R_{d_2}]\leq 1, \quad & \forall d_1,d_2 \in \mathcal{D}\\
&\nonumber x_{d_1}-z_{d_1} \leq \sum_{\substack{d_2 \in \mathcal{D} \\ R_{d_2}=R_{d_1}}}y_{d_1d_2}z_{d_2}, \quad & \forall d_1 \in \mathcal{D}\\ 
&\nonumber x_{d_1} \leq \sum_{r \in \hat{\mathcal{R}}}(R_{d_1r}+ \sum_{d_2 \in \mathcal{D}}y_{d_1d_2}z_{d_2}R_{d_2r}), & \quad \forall d_1 \in \mathcal{D}
\end{align}
\section{Column Generation Formulation of Multi-Person Pose Segmentation}
\label{solepose}
The LP relaxation of the ILP given in Eq  \ref{eq:MainProg} is very loose in practice and does not provide solutions that can be rounded to low cost integer solutions. Thus we propose to solve Eq \ref{eq:MainProg} via an alternative technique known as column generation\cite{barnprice}.  
Applying column generation results in an ILP. The corresponding LP relaxation is often integral in practice and when it is not integral, it is easily converted to a low cost integral solution.  The new ILP constructs poses using a set of variables corresponding to global and local poses which describe the global and local key-points respectively.  

In this alternative formulation, a global pose describes all global key-points in a common pose.  A local pose describes all key-points associated with a given part in a common pose.  
%
The sets of all possible global and local poses are exponentially large and thus can not be enumerated exactly.  However we can solve optimization over the global and local poses by relaxing the integrality constraints of the ILP and applying column generation to find a small sufficient set of global and local poses.  
%
%

%
\subsection{Global Poses}
\label{GlobalPose}
We define the set of all possible selections of global key-points in a single pose as $\mathcal{G}$ which we index with $q$. We refer to $\mathcal{G}$ as the set of global poses and its members as global poses.  Members of $\mathcal{G}$ must have at least one global key-point corresponding to a global part and no more than one key-point corresponding to any given part.  We describe  $\mathcal{G}$ using a matrix $G \in \{ 0,1\}^{|\mathcal{D}|\times |\mathcal{G}|}$. We set $G_{dq}=1$ if and only if key-point $d$ is associated with global pose $q$. 

We associate each $q\in \mathcal{G}$ with corresponding vectors $x^q\in \{0,1\}^{|\mathcal{D}|},y^q\in \{0,1\}^{|\mathcal{D}|\times |\mathcal{D}| },z^q \in \{0,1\}^{|\mathcal{D}|} $ as follows. 
\begin{align}
\nonumber &x^q_d=z^q_d=G_{dq}\quad &  \forall d \in \mathcal{D}\\
\nonumber &y^q_{d_1d_2}=G_{d_1q}G_{d_2q}=x^q_{d_1}x^q_{d_2} \quad & \forall d_1,d_2 \in \mathcal{D}
 \end{align}
We associate $\mathcal{G}$ with a cost vector $\Gamma \in \mathbb{R}^{|\mathcal{G}|}$ where $\Gamma_q$ is the cost associated with global pose $q$.  We define $\Gamma_q$ as follows.   
\begin{align}
\nonumber \Gamma_q&=\sum_{d\in \mathcal{D}}\theta_dx^q_d +\sum_{\substack{d_1\in \mathcal{D}\\d_2\in \mathcal{D}}}\phi_{d_1d_2}y^q_{d_1d_2}([R_{d_1}=R_{d_2}]+z^q_{d_1}z^q_{d_2})\\
 &=\sum_{d\in \mathcal{D}}\theta_dx^q_d +\sum_{d_1,d_2 \in \mathcal{D}}\phi_{d_1d_2}x^q_{d_1}x^q_{d_2}
\end{align}
\subsection{Local Poses}
\label{LocalPose}
We define the set of all possible selections of key-points corresponding to a shared part  in a common pose as $\mathcal{L}$ which we index with $q$. We refer to $\mathcal{L}$ as the set of local poses and its members as local poses.  Each member of $\mathcal{L}$ is associated with exactly one global key-point.  We describe $\mathcal{L}$  using $L,M \in \{0,1\}^{|\mathcal{D}| \times |\mathcal{L}|}$. Here $L_{dq}=1$ if and only if key-point $d$ is associated with $q$ as a local key-point.  Similarly $M_{dq}=1$ if and only if key-point $d$ is associated with $q$ as a global key-point.  
%
%
We associate each $q\in \mathcal{L}$ with corresponding vectors $x^q\in \{0,1\}^{|\mathcal{D}|},y^q\in \{0,1\}^{|\mathcal{D}|\times |\mathcal{D}| },z^q \in \{0,1\}^{|\mathcal{D}|} $ as follows. 
\begin{align}
\nonumber &x^q_d = L_{dq}+M_{dq}\quad& \forall d \in \mathcal{D}\\
\nonumber &y^q_{d_1d_2} = G_{d_1q}G_{d_2q}=x^q_{d_1}x^q_{d_2} \quad& \forall d_1,d_2 \in \mathcal{D}\\
\nonumber &z^q_d = M_{dq}\quad& \forall d \in \mathcal{D}
 \end{align}
We associate $\mathcal{L}$ with a cost vector $\Psi \in \mathbb{R}^{|\mathcal{L}|}$ where $\Psi_q$ is the cost associated with local pose $q$.  We define $\Psi_q$ as follows.   
\begin{align}
\nonumber \Psi_q&=\sum_{d\in \mathcal{D}}\theta_dx^q_d +\sum_{\substack{d_1\in \mathcal{D}\\d_2\in \mathcal{D}}}\phi_{d_1d_2}y^q_{d_1d_2}([R_{d_1}=R_{d_2}]+z^q_{d_1}z^q_{d_2})\\
 &= \sum_{d \in \mathcal{D}}\theta_d(x^q_d-z^q_d)+\sum_{d_1,d_2 \in \mathcal{D}}\phi_{d_1d_2}x^q_{d_1}x^q_{d_2}
\end{align}
\subsection{Objective and Constraints}
\label{objcon}
We define a selection of global and local poses using $\gamma \in \{0,1\}^{|\mathcal{G}|},\psi\in \{0,1\}^{|\mathcal{L}|}$ respectively.  We set $\gamma_q=1$ to indicate that global pose $q \in \mathcal{G}$ is selected and otherwise set $\gamma_q=0$. Similarly we set $\psi_q=1$ to indicate that local pose $q \in \mathcal{L}$ is selected and otherwise set $\psi_q=0$.
We define the cost associated with a selection of global and local poses as $\Gamma^t\gamma+\Psi^t\psi$. Minimizing the cost with respect to $\gamma,\psi$ does not result in a consistent poses since the variables $\gamma$,$\psi$ are related to each other. To ensure a consistent solution we enforce the following constraints:
\begin{itemize}
\item $G\gamma+L\psi \leq 1$: No key-point can be included as a global key-point in a global pose or local key-point in a local pose more than once.
\item $L\psi+M\psi \leq 1$: No key-point can be associated with more than one local pose.  
\item $-G\gamma+M\psi\leq 0 $:  Every local pose must associated with a global pose where the global key-point in the local pose is included in the corresponding global pose.
\end{itemize}
The objective and the above constraints form an ILP.
\begin{align}
\label{primdualilp}
\mbox{Eq }\ref{eq:MainProg}=\min_{\substack{\gamma \in \{0,1\}^{|\mathcal{G}|} \\ \psi \in \{0,1\}^{|\mathcal{L}|}}} & \Gamma^t\gamma+\Psi^t\psi \\
\text{s.t.} \quad \nonumber & G\gamma+L\psi \leq 1\\
\nonumber & L\psi+M\psi\leq 1\\
\nonumber & -G\gamma+M\psi\leq 0
\end{align}
\subsection{Primal and Dual}
\label{primDual}
By relaxing integral constraints on $\gamma,\psi$ we convert Eq \ref{primdualilp} to its dual form using Lagrange multiplier sets  $\lambda^1,\lambda^2,\lambda^3 \in \mathbb{R}_{0+}^{|\mathcal{D}|}$:  
\begin{align}
\label{dualFormPose}
\min_{\substack{\gamma\geq 0 \\ \psi \geq 0 \\ G\gamma+L\psi \leq 1 \\ L\psi+M\psi\leq 1 \\-G\gamma+M\psi\leq 0}}\Gamma^t\gamma+\Psi^t \psi 
=\max_{\substack{\lambda^1\geq 0\\ \lambda^2\geq 0 \\ \lambda^3\geq 0 \\  \Gamma+G^t(\lambda^1-\lambda^3) \geq 0 \\ \Psi+L^t\lambda^1+(M^t+L^t)\lambda^2+M^t\lambda^3 \geq 0}} -1^t\lambda^1-1^t\lambda^2-1^t\lambda^3 
\end{align}
\section{Generating Columns}
\label{violposes}
In this section we consider the problem of identifying violated constraints in the dual form in Eq \ref{dualFormPose}.  We divide this section into two parts.  
In Section \ref{mostviolloal} we study the production of the most violated constraint corresponding to a  local pose given that the global key-point in that pose (denoted $d_*$) is known.  By trying all $d \in \mathcal{D}$ for $d_*$ we are assured to find the most violated constraint corresponding to a  local pose.   

In Section \ref{mostviolglobal} we study the production of the most violated constraint corresponding to a global pose given that a single key-point ($d_*$) corresponding to a global part is included.  By trying all possible $d_* \in \mathcal{D} \quad \text{s.t.} \quad R_{d_*} \in \hat{\mathcal{R}}$ we are assured to find the most violated constraint corresponding to a global pose.  
\subsection{Violated Local Poses}
\label{mostviolloal}
For each $d_*\in \mathcal{D}$ we compute the most violated constraint corresponding to a local pose in which $d_*$ is the global key-point.  We write this as an ILP below.
\begin{align}
\label{mostViolGlobalEqu}
\min_{\substack{q\in \mathcal{G} \\ M_{d_*q}=1}}&(\lambda^2_{d_*}+\lambda^3_{d_*})M_{d_*q}+\sum_{d \in \mathcal{D}}(\lambda^1_d+\lambda^2_d)L_{dq}+\Psi_q \\
\nonumber =\min_{\substack{x \in \{ 0,1\}^{|\mathcal{D}|} \\x_{d_*}=1}}&(-\lambda^1_{d_*}+\lambda^3_{d_*}-\theta_{d_*}) +\sum_{d \in \mathcal{D}}(\theta_d+\lambda^1_{d}+\lambda^2_d)x_d
\nonumber +\sum_{d_1,d_2 \in \mathcal{D}}x_{d_1}x_{d_2}\phi_{d_1d_2}
\end{align}
With proper thresholding, the number of key-points associated with any given part in an image is small.  It is no more than 15 in practice and generally less than ten.  Thus solving the ILP above is feasible. 
\subsection{Violated Global Poses}
\label{mostviolglobal}
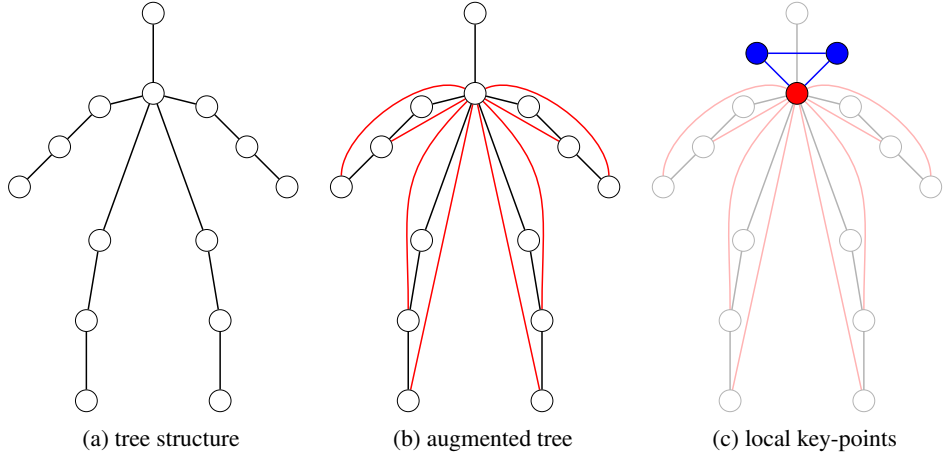
\begin{figure}[t]
\begin{center}
\begin{subfigure}[b]{0.3\textwidth}
\resizebox{0.95\textwidth}{!}{
\begin{tikzpicture}
	\begin{pgfonlayer}{nodelayer}
		\node [style=vertex_default] (0) at (-3.75, 5) {};      
		\node [style=vertex_default] (1) at (-3.75, 3.5) {};    
		\node [style=vertex_default] (2) at (-4.75, 3.25) {};   
		\node [style=vertex_default] (3) at (-2.75, 3.25) {};   
		\node [style=vertex_default] (4) at (-5.5, 2.5) {};     
		\node [style=vertex_default] (5) at (-2, 2.5) {};       
		\node [style=vertex_default] (6) at (-6.25, 1.75) {};   
		\node [style=vertex_default] (7) at (-1.25, 1.75) {};   
		\node [style=vertex_default] (8) at (-4.75, 0.75) {};   
		\node [style=vertex_default] (9) at (-2.75, 0.75) {};   
		\node [style=vertex_default] (10) at (-5, -0.75) {};    
		\node [style=vertex_default] (11) at (-2.5, -0.75) {};  
		\node [style=vertex_default] (12) at (-5, -2.25) {};    
		\node [style=vertex_default] (13) at (-2.5, -2.25) {};  
	\end{pgfonlayer}
	\begin{pgfonlayer}{edgelayer}
		\draw [style=tree_edge] (0) to (1);
		\draw [style=tree_edge] (1) to (2);
		\draw [style=tree_edge] (1) to (3);
		\draw [style=tree_edge] (2) to (4);
		\draw [style=tree_edge] (3) to (5);
		\draw [style=tree_edge] (4) to (6);
		\draw [style=tree_edge] (5) to (7);
		\draw [style=tree_edge] (1) to (8);
		\draw [style=tree_edge] (1) to (9);
		\draw [style=tree_edge] (8) to (10);
		\draw [style=tree_edge] (9) to (11);
		\draw [style=tree_edge] (10) to (12);
		\draw [style=tree_edge] (11) to (13);
	\end{pgfonlayer}
\end{tikzpicture}
}
\caption{tree structure}
\end{subfigure}
\begin{subfigure}[b]{0.3\textwidth}
\resizebox{0.95\textwidth}{!}{
\begin{tikzpicture}
	\begin{pgfonlayer}{nodelayer}
		\node [style=vertex_default] (0) at (-3.75, 5) {};      
		\node [style=vertex_default] (1) at (-3.75, 3.5) {};    
		\node [style=vertex_default] (2) at (-4.75, 3.25) {};   
		\node [style=vertex_default] (3) at (-2.75, 3.25) {};   
		\node [style=vertex_default] (4) at (-5.5, 2.5) {};     
		\node [style=vertex_default] (5) at (-2, 2.5) {};       
		\node [style=vertex_default] (6) at (-6.25, 1.75) {};   
		\node [style=vertex_default] (7) at (-1.25, 1.75) {};   
		\node [style=vertex_default] (8) at (-4.75, 0.75) {};   
		\node [style=vertex_default] (9) at (-2.75, 0.75) {};   
		\node [style=vertex_default] (10) at (-5, -0.75) {};    
		\node [style=vertex_default] (11) at (-2.5, -0.75) {};  
		\node [style=vertex_default] (12) at (-5, -2.25) {};    
		\node [style=vertex_default] (13) at (-2.5, -2.25) {};  
	\end{pgfonlayer}
	\begin{pgfonlayer}{edgelayer}
		\draw [style=tree_edge] (0) to (1);
		\draw [style=tree_edge] (1) to (2);
		\draw [style=tree_edge] (1) to (3);
		\draw [style=tree_edge] (2) to (4);
		\draw [style=tree_edge] (3) to (5);
		\draw [style=tree_edge] (4) to (6);
		\draw [style=tree_edge] (5) to (7);
		\draw [style=tree_edge] (1) to (8);
		\draw [style=tree_edge] (1) to (9);
		\draw [style=tree_edge] (8) to (10);
		\draw [style=tree_edge] (9) to (11);
		\draw [style=tree_edge] (10) to (12);
		\draw [style=tree_edge] (11) to (13);

		\draw [style=add_edge] (1) to (4);
		\draw [style=add_edge] (1) to (5);
		\draw [style=add_edge, in=90, out=155, looseness=.75] (1) to (6);
		\draw [style=add_edge, in=90, out=25, looseness=.75] (1) to (7);
		\draw [style=add_edge, in=90, out=225, looseness=1.25] (1) to (10);
		\draw [style=add_edge, in=90, out=315, looseness=1.25] (1) to (11);
		\draw [style=add_edge] (1) to (12);
		\draw [style=add_edge] (1) to (13);
	\end{pgfonlayer}
\end{tikzpicture}
}
\caption{augmented tree}
\end{subfigure}
\begin{subfigure}[b]{0.3\textwidth}
\resizebox{0.95\textwidth}{!}{
\begin{tikzpicture}
	\begin{pgfonlayer}{nodelayer}
		\node [style=vertex_default, draw opacity=.3] (0) at (-3.75, 5) {};      
		\node [style=vertex_default, fill=Red] (1) at (-3.75, 3.5) {};    
		\node [style=vertex_default, draw opacity=.3] (2) at (-4.75, 3.25) {};   
		\node [style=vertex_default, draw opacity=.3] (3) at (-2.75, 3.25) {};   
		\node [style=vertex_default, draw opacity=.3] (4) at (-5.5, 2.5) {};     
		\node [style=vertex_default, draw opacity=.3] (5) at (-2, 2.5) {};       
		\node [style=vertex_default, draw opacity=.3] (6) at (-6.25, 1.75) {};   
		\node [style=vertex_default, draw opacity=.3] (7) at (-1.25, 1.75) {};   
		\node [style=vertex_default, draw opacity=.3] (8) at (-4.75, 0.75) {};   
		\node [style=vertex_default, draw opacity=.3] (9) at (-2.75, 0.75) {};   
		\node [style=vertex_default, draw opacity=.3] (10) at (-5, -0.75) {};    
		\node [style=vertex_default, draw opacity=.3] (11) at (-2.5, -0.75) {};  
		\node [style=vertex_default, draw opacity=.3] (12) at (-5, -2.25) {};    
		\node [style=vertex_default, draw opacity=.3] (13) at (-2.5, -2.25) {};  
		\node [style=vertex_default, fill=Blue] (14) at (-4.5, 4.25) {};    
		\node [style=vertex_default, fill=Blue] (15) at (-3, 4.25) {};  
	\end{pgfonlayer}
	\begin{pgfonlayer}{edgelayer}
		\draw [style=tree_edge, draw opacity=.3] (0) to (1);
		\draw [style=tree_edge, draw opacity=.3] (1) to (2);
		\draw [style=tree_edge, draw opacity=.3] (1) to (3);
		\draw [style=tree_edge, draw opacity=.3] (2) to (4);
		\draw [style=tree_edge, draw opacity=.3] (3) to (5);
		\draw [style=tree_edge, draw opacity=.3] (4) to (6);
		\draw [style=tree_edge, draw opacity=.3] (5) to (7);
		\draw [style=tree_edge, draw opacity=.3] (1) to (8);
		\draw [style=tree_edge, draw opacity=.3] (1) to (9);
		\draw [style=tree_edge, draw opacity=.3] (8) to (10);
		\draw [style=tree_edge, draw opacity=.3] (9) to (11);
		\draw [style=tree_edge, draw opacity=.3] (10) to (12);
		\draw [style=tree_edge, draw opacity=.3] (11) to (13);

		\draw [style=add_edge, draw opacity=.3] (1) to (4);
		\draw [style=add_edge, draw opacity=.3] (1) to (5);
		\draw [style=add_edge, in=90, out=155, looseness=.75, draw opacity=.3] (1) to (6);
		\draw [style=add_edge, in=90, out=25, looseness=.75, draw opacity=.3] (1) to (7);
		\draw [style=add_edge, in=90, out=225, looseness=1.25, draw opacity=.3] (1) to (10);
		\draw [style=add_edge, in=90, out=315, looseness=1.25, draw opacity=.3] (1) to (11);
		\draw [style=add_edge, draw opacity=.3] (1) to (12);
		\draw [style=add_edge, draw opacity=.3] (1) to (13);

		\draw [style=local_edge] (1) to (14);
		\draw [style=local_edge] (1) to (15);
		\draw [style=local_edge] (14) to (15);
	\end{pgfonlayer}
\end{tikzpicture}
}
\caption{local key-points}
\end{subfigure}
\end{center}
\caption{Graphical representation of tree-structured model and its extension
  with additional pairwise terms and local key-points. (a) Traditional pictorial
  structure that represents articulated poses for human. In our formulation we
  use 14 parts: head, neck, left/right shoulder, left/right elbow, left/right
  wrist, left/right hip, left/right kneel and left/right ankle. (b) Tree
  structure augmented with additional pairwise terms from neck to all other
  non-adjacent parts of human body (red edges). (c) Each body part can be
  associated with multiple key-points, among which there can be only one global
  key-point (red node) and multiple local key-point (blue nodes); key-points
associated with the same part of the same person forms a fully-connected
sub-graph. Local key-points of different parts of the same person do not
interact with each other directly.}
\label{fig:tree_model}
\end{figure}

For each $d_*$ such that $R_{d_*} \in \hat{\mathcal{R}}$ we compute the most violated constraint corresponding to a global pose including $d_*$.  We write this as an integer program below.
\begin{align}
\min_{\substack{q \in \mathcal{G} \\ G_{d_*q}=1}}&\Gamma_q+\sum_{d \in \mathcal{D}} G_{dq}(\lambda^1_d-\lambda^3_d)\\
\nonumber =\min_{ \substack{z \in \{ 0,1\}^{|\mathcal{D}|} \\ z_{d_*}=1\\  \sum_{d \in \mathcal{D}}R_{dr}z_d \leq 1 \; \; \forall r \in \mathcal{R}}}&\sum_{ d\in \mathcal{D}} (\theta_{d}+\lambda^1_d-\lambda^3_d)z_{d}
\nonumber +\sum_{d_1d_2 \in \mathcal{D}} \phi_{d_1d_2}z_{d_1}z_{d_2}
\end{align}
%
Different types of structure on $\phi$ can result in easier inference.  A typical model is the tree structure over parts of the human body as in the deformable part model of \cite{deva1,deva2,deva3}.  $\phi$ will be zero between non-adjacent parts on the tree.  We augment this tree model with additional edges from neck to all other non-adjacent body parts to handle situations where certain parts are occluded/not visible. This is illustrated in Fig~\ref{fig:tree_model}.  

Given a tree structured model exact inference can be achieved via dynamic programming.  In our application we strengthen the tree model by connecting the neck to every other part and making the neck the only global part. Given that the global key-point associated with the neck is fixed, then exact inference can be achieved via dynamic programming.  Thus Eq \ref{mostViolGlobalEqu} can be solved by solving one dynamic program for each possible choice of key-point associated with the neck.  
%
%

%
\subsection{Column Generation Algorithm}
\label{colgenalg}
We write our column generation algorithm in Alg \ref{dualsolvesimpless} and describe it below.  We construct nascent subsets of $\mathcal{G},\mathcal{L}$ which we denote as $\hat{\mathcal{G}},\hat{\mathcal{L}}$.  
At each step we compute the most violated constraint corresponding to a local pose in which key-point $d$ is the global key-point for all $d\in \mathcal{D}$.  
At each step we also compute the most violated constraint corresponding to a global pose in which key-point $d$ is included for all $d\in \mathcal{D}$ such that $R_d\in \hat{\mathcal{R}}$. 

Poses are added to the nascent sets $\hat{\mathcal{G}},\hat{\mathcal{L}}$ if and only if the constraint is violated.  
We terminate when no violated constraints exist and return the primal solution.  
\begin{algorithm}[H]
    \caption{Dual Optimization }
\begin{algorithmic} 
\State $\hat{G} \leftarrow \{ \}$
\State $\hat{L} \leftarrow \{ \}$
\Repeat
\State $\lambda \leftarrow$ Maximize dual in Eq \ref{dualFormPose} over column sets $\hat{\mathcal{G}},\hat{\mathcal{L}}$
\For {$d_* \in \mathcal{D}$}
\State $q_* \leftarrow \mbox{arg} \min_{\substack{q\in \mathcal{G} \\ M_{d_*q}=1}}(\lambda^3_{d_*}+\lambda^2_{d_*})M_{d_*q}+\sum_{d \in \mathcal{D}}(\lambda^1_d+\lambda^2_d)L_{dq}+\Psi_q$
\If{$(\lambda^3_{d_*}+\lambda^2_{d_*})M_{d_*q_*}+\sum_{d \in \mathcal{D}}(\lambda^1_d+\lambda^2_d)L_{dq_*}+\Psi_{q_*}<0$}
\State $\dot{\mathcal{L}} \leftarrow [\dot{\mathcal{L}} \cup q_*]$
\EndIf
\EndFor
\For {$d_{*} \in \mathcal{D}$  s.t. $ R_{d_*}$  $ \in \hat{\mathcal{R}}}$
\State $q_* \leftarrow \mbox{arg} \min_{\substack{q \in \mathcal{G} \\ G_{d_*q}=1}}\Gamma_q+\sum_{d \in \mathcal{D}} G_{dq}(\lambda^1_d-\lambda^3_d)$
\If{$\Gamma_{q_*}+\sum_{d \in \mathcal{D}} G_{dq_*}(\lambda^1_d-\lambda^3_d)<0$}
\State $\dot{\mathcal{G}} \leftarrow [\dot{\mathcal{G}} \cup q_*]$
\EndIf
\EndFor
\State  $\hat{\mathcal{L}}\leftarrow [\hat{\mathcal{L}},\dot{\mathcal{L}}]$
\State  $\hat{\mathcal{G}}\leftarrow [\hat{\mathcal{G}},\dot{\mathcal{G}}]$
%
 \Until{ $|\dot{\mathcal{G}}|+|\dot{\mathcal{L}}|  =0 $}
\State Recover $\gamma$ from $\lambda$.  NOTE:  $\gamma $ is computed for free by the LP solver when solving for $\lambda$.%
\end{algorithmic}
\label{dualsolvesimpless}
  \end{algorithm}
\section{Formulating Instance Segmentation for Column Generation }
\label{Cells}
%
%
We now consider our approach for densely clustered cell instance segmentation.  Given an image we first compute a set of super-pixels which provides an over-segmentation of the image.  We then construct an optimization problem that groups the super-pixels into small coherent cells or labels them as background. 
This optimization problem is an ILP.    

We outline our approach to instance segmentation as follows.  In Section \ref{cellSegCol} we describe the set of possible cells  and associate a cost to each according to a model.  In Section \ref{optformcell} we formulate instance segmentation as an ILP that is amenable to  attack using column generation.  We then relax the integrality and take the dual.  In Section \ref{colbb} we study the production of columns via identifying violated constraints in the dual.  In Section \ref{algcell} we detail our instance segmentation algorithm.  
\subsection{Cell Segmentation as Column Generation }
\label{cellSegCol}

We denote the set of all possible cells by $\mathcal{Q}$ and use $ G \in \{ 0,1\}^{
|\mathcal{D}|\times|\mathcal{Q}|}$ to denote the super-pixel-cell incidence matrix where $G_{dq}=1$ if and only if cell $q$ includes super-pixel $d$. Cells are constrained to have a maximum radius (denoted $m_R$) and maximum area/volume (denoted $m_V$).  Here $m_V$ and $m_R$ are model defined parameters.  We describe the distance between pairs of super-pixels, indexed with $d_1,d_2$, using a matrix $S\in \mathbb{R}_{0+}^{|\mathcal{D}|\times|\mathcal{D}|}$ where the distance between the centers of super-pixels $d_1,d_2$  is denoted  $S_{d_1d_2}$.  Given $S$ we describe the constraint on the radius of a cell as
  \begin{align}
  \label{centroid}
 \exists[d_*;
 G_{d_*q}=1] \quad \mbox{ s.t. }\;  0=\sum_{d_2 \in \mathcal{D}}G_{d_2q}[S_{d_*,d_2}\geq m_R] \quad \forall q \in \mathcal{Q}.
 \end{align}
For any given $q \in \mathcal{Q}$ any argument $d_*\in \mathcal{D}$ satisfying Eq~\ref{centroid} is called a centroid of $q$.  We now describe the constraint on the area of a cell.  We use $V \in \mathbb{R}_{+}^{|\mathcal{D}|}$ to describe the area of super-pixels which we index with $d$.  Here $V_d$ denotes the area of a particular super-pixel $d$.  Given $V$ we write the constraint on the area of a cell as  
 \begin{align}
 m_V\geq \sum_{d \in \mathcal{D}}G_{dq}V_d \quad \forall q \in \mathcal{Q}.
 \end{align}
We denote the costs associated with cells as $\Gamma \in \mathbb{R}^{|\mathcal{Q}|}$, where $\Gamma_q$ describes the cost associated with cell $q$:
\begin{align}
\Gamma_q=\sum_{d \in \mathcal{D}}\theta_{d}G_{dq}+\sum_{\substack{d_1\in \mathcal{D} \\ d_2\in \mathcal{D}}}\phi_{d_1d_2}G_{d_1q}G_{d_2q}. 
\end{align}
%
We use $\theta_d$ to denote the cost  associated with including $d$ in a cell.  Similarly we use $\phi \in \mathbb{R}^{|\mathcal{D}| \times |\mathcal{D}|}$, where $\phi_{d_1d_2}$ denotes the cost associated with including $d_1$ and $d_2$ in a common cell.  

\subsection{Optimization Formulation}
\label{optformcell}
A solution to instance segmentation is
denoted by the indicator vector $\gamma  \in \{0,1\}^{|\mathcal{Q}|}$ where
$\gamma_q=1$ indicates that cell $q$ is included in the solution and $\gamma_q=0$ otherwise.  A collection of cells specified by $\gamma$ is a valid solution if and only if each super-pixel is associated with at most one active cell.  We now express instance segmentation as an ILP.
\begin{align}
\label{intobjCell}
\min_{ \substack{\gamma \in \{0,1\}^{|\mathcal{Q}|}\\G\gamma \leq 1 }}\Gamma^t \gamma
\end{align}
We now attack optimization in Eq \ref{intobjCell} using the well studied tools of LP relaxations. We write the primal and dual LP relaxation below using Lagrange multipliers $\lambda \in \mathbb{R}_{0+}^{|\mathcal{D}|}$.
\begin{align}
\label{lpcell}
\mbox{Eq } \ref{intobjCell} \geq \quad \min_{\substack{ \gamma \geq 0 \\G\gamma \leq 1}}\Gamma^t \gamma=\max_{\substack{\lambda \geq 0\\ \Gamma+G^t\lambda \geq 0}}-1^t\lambda 
\end{align}
Even under conservative constraints on the radius and area of a cell, the number or primal variables in Eq \ref{lpcell} is intractable to enumerate much during the optimization. However the number of primal constraints is equal to the number of super-pixels which is small, i.e. a couple of thousands. This motivates the use of column generation.  
\subsection{Generating Columns}
\label{colbb}
We now consider the problem of finding violated constraints in the dual of Eq \ref{lpcell}.  We do this by solving one ILP for each $d_* \in \mathcal{D}$.  This optimization computes the most violated constraint corresponding to a cell such that $G_{d_*q}=1$ and in which $d_*$ is a centroid.  We write the optimization below.  We describe the solution using indicator vector $x\in \{ 0,1\}^{|\mathcal{D}}$. 
\begin{align}
\label{getcolcell}
\min_{\substack{q \in \mathcal{Q} \\ G_{d_*q}=1}}\Gamma_q+\sum_{d \in \mathcal{D}}G_{dq}\lambda_{d} 
 =\min_{\substack{x \in \{ 0,1\}^{|\mathcal{Q}|} \\ x_{d_*}=1 \\  x_d =0 \; ;\;S_{d,d_*}>m_R\\ \sum_{d\in \mathcal{D}}x_d \leq m_V}} \sum_{d \in \mathcal{D}}(\theta_d+\lambda_dx_d)+\sum_{d_1,d_2 \in \mathcal{D}}\phi_{d_1d_2}x_{d_1}x_{d_2}
\end{align}
Given  the solution $x$ the most violated column $q$ corresponding to a cell with centroid $d_*$ is defined as $G_{dq}=x_d$ for all $d \in \mathcal{D}$.  In our data sets the maximum radius of a cell is rather small.  Thus the selection of the centroid reduces the total number of other super-pixels that need be considered to less than twenty.  Thus attacking optimization in Eq \ref{getcolcell} with an ILP  is very efficient.   Furthermore solving the ILP corresponding to each $d_* \in \mathcal{D}$ can be done in parallel.  
\subsection{Column Generation Algorithm for Cell Segmentation}
\label{algcell}
We write our column generation algorithm in Alg \ref{cellalg} and describe it below.  We construct a nascent subset of $\mathcal{Q}$ which we denote as $\hat{\mathcal{Q}}$.  
At each step we compute the most violated constraint corresponding to a cell in which a centroid is  super-pixel $d$, for all $d\in \mathcal{D}$.  We add this to $\mathcal{Q}$ if and only if it corresponds to a violated constraint. 
\begin{algorithm}[H]
    \caption{Dual Optimization }
\begin{algorithmic} 
\State $\hat{Q} \leftarrow \{ \}$
\Repeat
\State $\lambda \leftarrow$ Maximize dual in Eq \ref{lpcell} over constraint set $\hat{\mathcal{Q}}$
\For {$d_* \in \mathcal{D}$}
\State $q_* \leftarrow \mbox{arg} \min_{\substack{q \in \mathcal{Q} \\ G_{d_*q}=1}}\Gamma_q+\sum_{d \in \mathcal{D}}G_{dq}\lambda_{d} $
\If{$\Gamma_{q_*}+\sum_{d \in \mathcal{D}}G_{dq_*}\lambda_{d}<0$}
\State $\dot{\mathcal{Q}} \leftarrow [\dot{\mathcal{Q}} \cup q_*]$
\EndIf
\EndFor
\State  $\hat{\mathcal{Q}}\leftarrow [\hat{\mathcal{Q}},\dot{\mathcal{Q}}]$
%
 \Until{ $|\dot{\mathcal{Q}}|=0 $}
\State Recover $\gamma$ from $\lambda$.  NOTE:  $\gamma $ is computed for free by the LP solver when solving for $\lambda$.%
\end{algorithmic}
\label{cellalg}
 \end{algorithm}
\section{Anytime Bounds}
\label{lbUb}
In this section we discuss the construction of anytime upper and lower bounds on the optimal integer solution to our problems.  Every upper bound is associated with an integer solution.  
\subsection{Upper Bounds}
We found that in practice our LP relaxations are integral at termination and generally integral after each step of optimization.  
However in cases where the LP is loose we found that solving the ILP given the primal variables generated took negligible additional time beyond solving the LP for pose segmentation and little additional time for cell segmentation.  However we can use various rounding procedures such as that of \cite{yarkoNips2016} if difficult ILPs occur. For the case of cell segmentation we write the rounding procedure of \cite{yarkoNips2016} below.  
\subsubsection{Rounding Fractional Solutions}
\label{round}
We attack rounding a
fractional $\gamma$ via a greedy iterative approach that at each
iteration, selects the cell $q$ with minimum value $\Gamma_q\gamma_q$ discounted by the
fractional cost of any cells that share a superpixel with $q$ (and hence can no
longer be added to the segmentation if $q$ is added).  We write the rounding
procedure in Alg \ref{ubr} using the notation $\mathcal{Q}_{\perp q}$ to indicate
the set of cells in $\mathcal{Q}$ that intersect cell $q$ (excluding $q$
itself).
 \begin{algorithm}[H]
    \caption{Upper Bound Rounding}
\begin{algorithmic} 
\While {$\exists q \in \mathcal{Q} \quad \mbox{ s.t. } \gamma_q \notin \{ 0,1 \}$}
\State $q^*\leftarrow \mbox{arg}\min_{\substack{q \in \mathcal{Q} \\ \gamma_q>0}}\Gamma_q\gamma_q -\sum_{\hat{q} \in \mathcal{Q}_{\perp q}} \gamma_{\hat{q}}\Gamma_{\hat{q}} $ \\
\State $\gamma_{\hat{q}} \leftarrow 0 \quad \forall \hat{q} \in \mathcal{Q}_{\perp q^*}$
\State $\gamma_{q^*}\leftarrow 1 $
\EndWhile
\State RETURN $\gamma$
\end{algorithmic}
\label{ubr}
  \end{algorithm}
\subsection{Anytime Lower Bounds}
\label{lowerplacmain}
%
%
Computing an anytime lower bound is done using the value of the most violated constraint for each problem solved when generating columns at each itteration.  All non-positive values are then summed and added to the value of the LP relaxation to produce a lower bound.  This lower bound has value equal to the LP at convergence of column generation since at termination no violated constraints exist. We write the computation of the bounds for the case of pose and cell segmentation below.  They are valid for any non-negative setting of the dual variables .  We derive both bounds in Appendix \ref{explower}.  
 \begin{align}
\mbox{Eq }\ref{primdualilp} \geq &-1^t \lambda^1-1^t\lambda^2-1^t\lambda^3\\
&\nonumber +\sum_{d_* \in \mathcal{D}}\min[0,\min_{\substack{q\in \mathcal{G} \\ M_{d_*q}=1}}(\lambda^3_{d_*}-\lambda^1_{d_*})M_{d_*q}+\sum_{d \in \mathcal{D}}(\lambda^1_d+\lambda^2_d)L_{dq}+\Psi_q ]\\
&\nonumber +\sum_{\substack{d_* \in \mathcal{D}\\ R_{d_*} \in \hat{\mathcal{R}}}}\min[0,\min_{\substack{q \in \mathcal{G} \\ G_{d_*q}=1}}\Gamma_q+\sum_{d \in \mathcal{D}} G_{dq}(\lambda^1_d-\lambda^3_d)]
\end{align}
\begin{align}
 \label{lowerformdoc}
\mbox{Eq }\ref{intobjCell}  \geq -1^t\lambda +\sum_{d_* \in \mathcal{D}}\min [ 0,  \min_{\substack{q \in \mathcal{Q} \\ G_{d_*q}=1}}\Gamma_q+\sum_{d \in \mathcal{D}}G_{dq}\lambda_{d}]
 \end{align}
\section{Experiments}
\label{poseCellExp}
\subsection{Multi-Person Pose Segmentation}
\label{ssec:pose}
We evaluate our approach in terms of mean Average Precision (mAP) on MPII-Multiperson training set~\cite{andriluka14cvpr} which consists of 3844 images. We directly use the unary and pairwise potentials provided by the authors of~\cite{deepcut1}. We compare our results against that of~\cite{deepcut1}, which are also provided by the authors. 

We offset $\Gamma$ with a constant set heuristically at value thirty to discourage the selection of global poses including few key-points, which tend to be lower magnitude in cost.  We tighten our relaxation using odd set inequalities \cite{heismann2014generalization} of size three as done \cite{yarkoNips2016} which are discussed in Appendix \ref{betterLP} though these are rarely needed.  We also bound $\lambda$ as discussed in Appendix \ref{bounPoseBase} though we found this to make little difference in practice.

Our approach runs much faster that the baseline due to the reduced model size and a more sophisticated inference algorithm, while we also perform slightly better in term of mean average precision.  

\begin{figure*}[t]
\begin{center}
\begin{tabular}{ccc}
\includegraphics[width=0.325\textwidth]{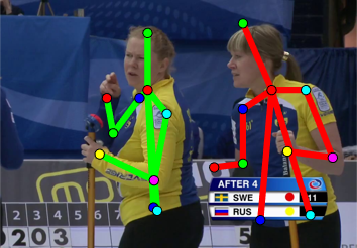} &
\includegraphics[width=0.31\textwidth]{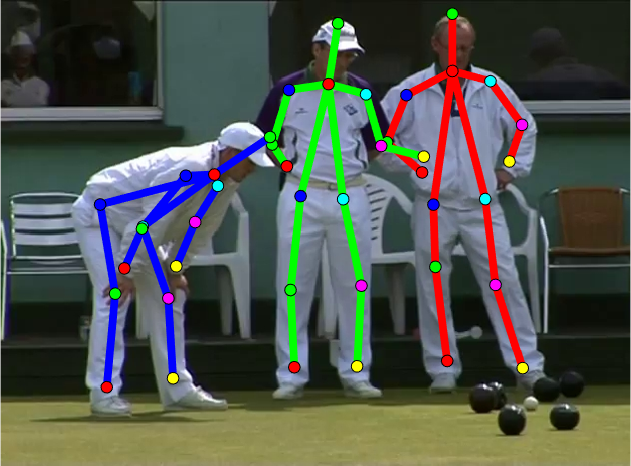} &
\includegraphics[width=0.29\textwidth]{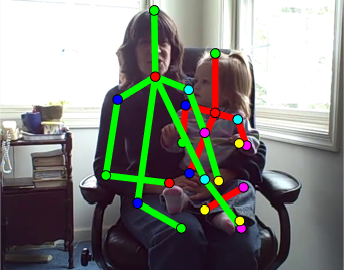}
\end{tabular}
\caption{Sample image produced with out approach.  Different people are annotated with specific colors and different body parts with colored dots.
}
\label{samplePoseIm}
\end{center}
\end{figure*}

\begin{figure}[t]
\begin{center}
\begin{tabular}{ c | c c c}
\rotatebox[origin=c]{90}{Deeper Cut \cite{deepcut1}} &
\raisebox{-.5\height}{\includegraphics[width=0.275\textwidth]{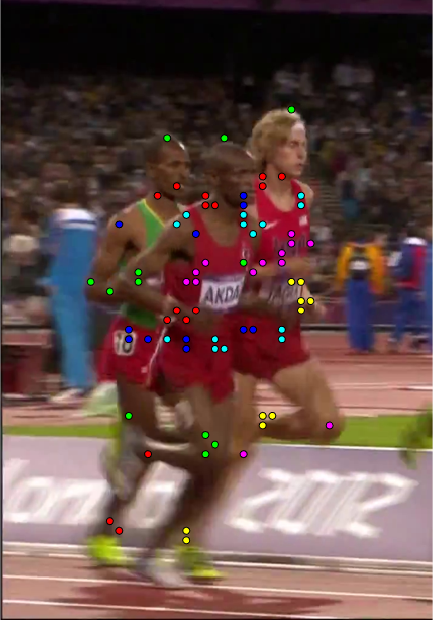}} &
\raisebox{-.5\height}{\includegraphics[width=0.275\textwidth]{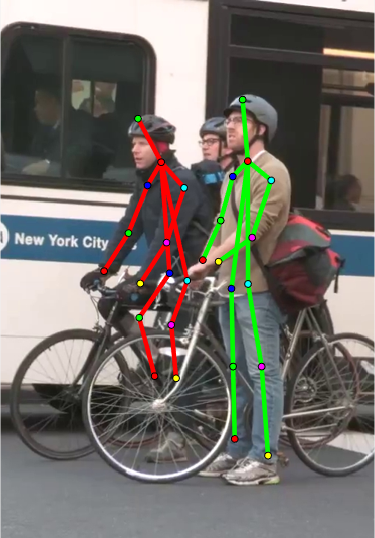}} &
\raisebox{-.5\height}{\includegraphics[width=0.2727\textwidth]{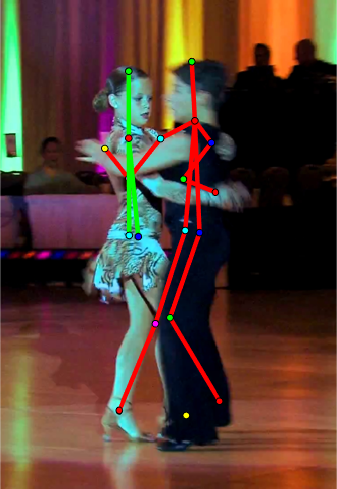}} \\
\rotatebox[origin=c]{90}{Our Approach} &
\raisebox{-.5\height}{\includegraphics[width=0.275\textwidth]{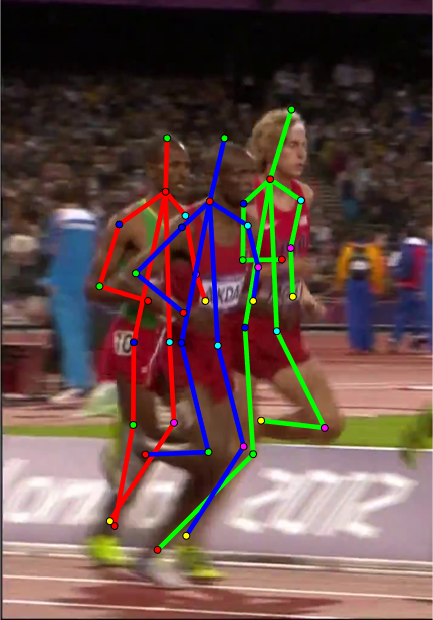}} &
\raisebox{-.5\height}{\includegraphics[width=0.275\textwidth]{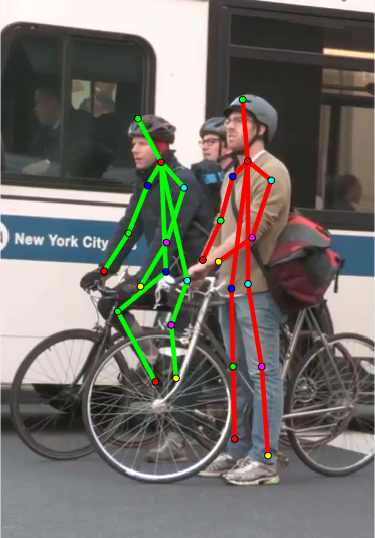}} &
\raisebox{-.5\height}{\includegraphics[width=0.2727\textwidth]{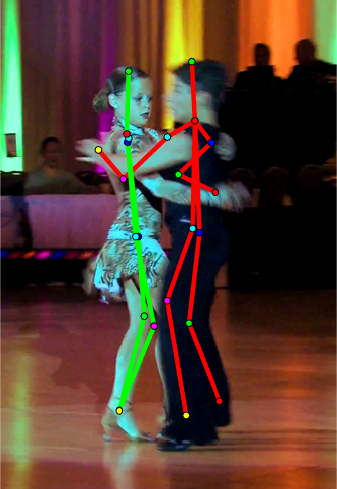}}
\end{tabular}
\caption{Qualitative comparison of \cite{deepcut1} (\emph{top row}) and our approach (\emph{bottom row}).
(\emph{Left column}) \cite{deepcut1} occasionally fails and produces many false positives per detection,
while our approach avoid this by enforcing the fact that each individual person must have a neck. (\emph{Middle column})
We predict right kneel of the person on the left better than \cite{deepcut1}. (\emph{Right column}) \cite{deepcut1}
fails to find the lower body parts of the person on the left and confuses ankle and kneel of the two people, while we
successfully avoid these errors. }
\label{figgood333}
\end{center}
\end{figure}

\begin{figure}
\begin{center}
\begin{tabular}{cc}
\includegraphics[clip,trim=.5cm 7cm 3cm 7cm,width=0.45\textwidth]{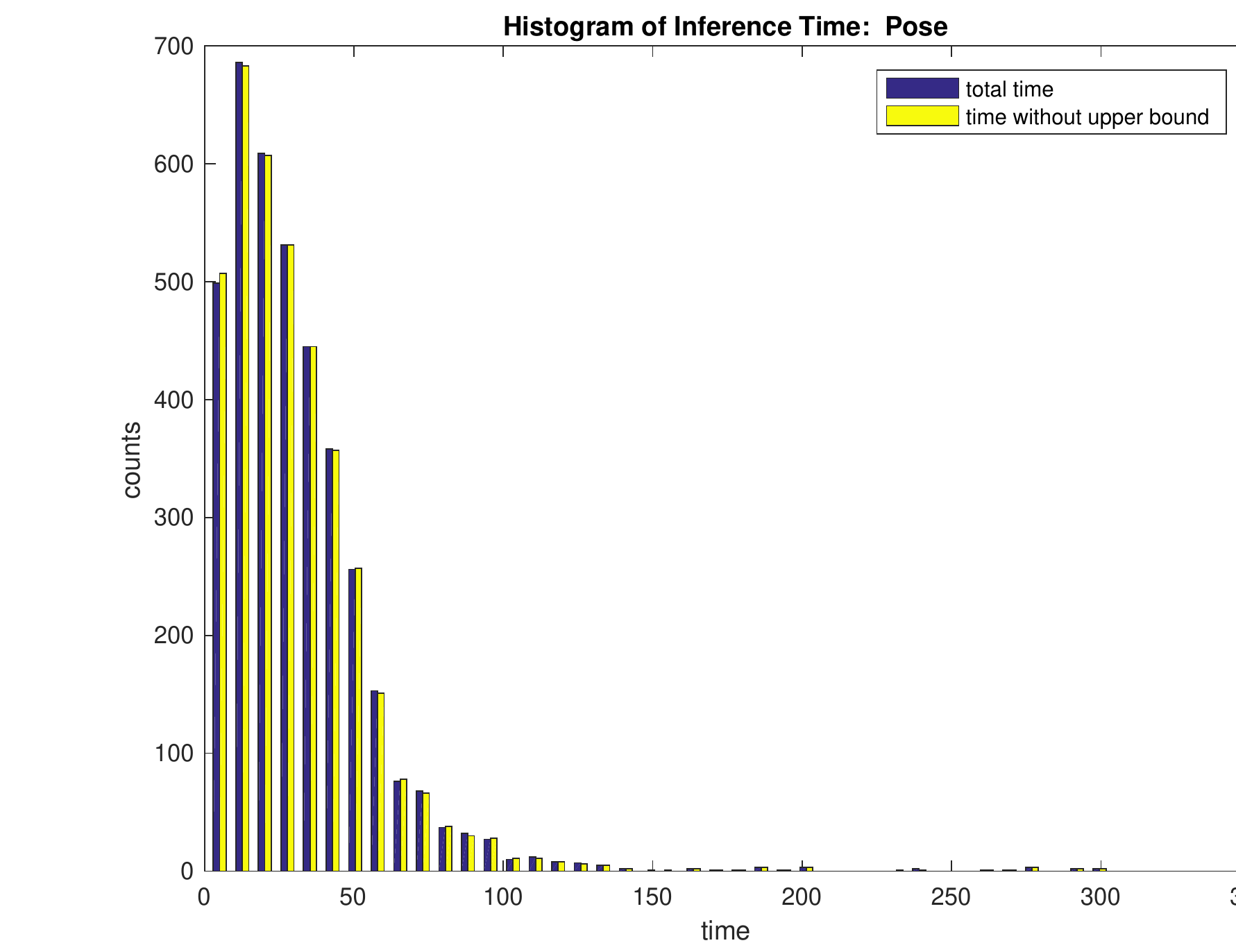}
\includegraphics[clip,trim=.5cm 7cm 3cm 7cm,width=0.45\textwidth]{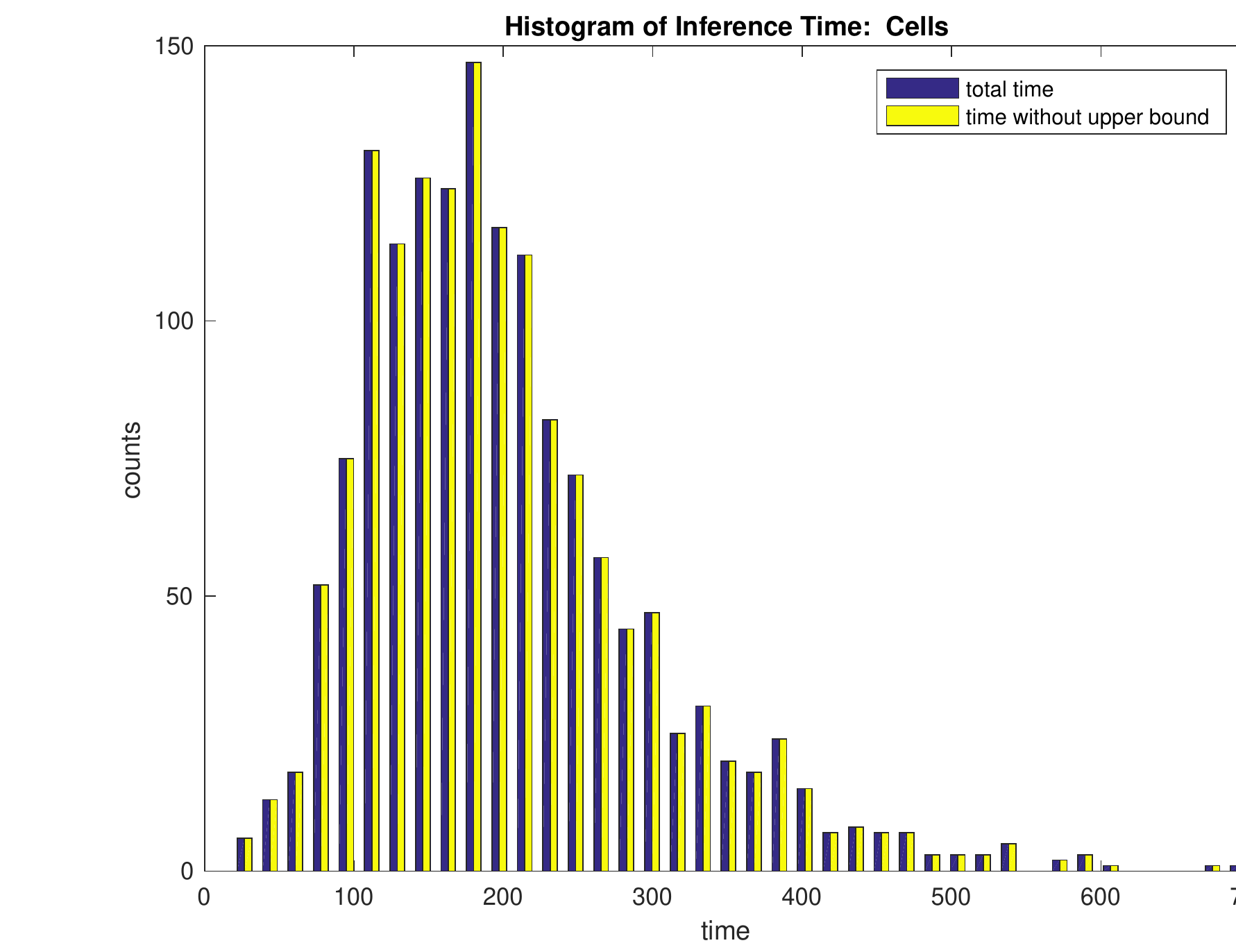}
\end{tabular}
\caption{\textbf{Left}:  Histogram of inference time for pose segmentation; \textbf{Right}:  Histogram of inference time for cell segmentation 
} 
\label{histplotpose}
\end{center}
\end{figure}

We display a sample result in Fig \ref{samplePoseIm} and performance on benchmarks in Table~\ref{pose-estimation}.  In Figure \ref{figgood333} we add qualitative comparisons to supplement our quantitative comparisons. We plot a histogram of inference time in Fig \ref{histplotpose}. 

We now consider the difference  between the upper and lower bounds at termination which we refer to as the gap. We divide the gap by the lower bound  and take the absolute value to normalize the gap.  
We found that  98.5\% out of 3844 images had a normalized gap of zero and hence were solved exactly. We observe that (100,99.95,99.40,98.888,98.660)\% of problem instances had a normalized gap under (0.16,0.1,0.01,0.001,0.0001) respectively.  
\begin{table*}[t]
\begin{center}
 \begin{tabular}{||c c c c c c c c c c | c||} 
 \hline
 Part & Head & Shoulder & Elbow & Wrist  & Hip & Knee& Ankle & UBody & Total & time (s/frame)\\ [0.5ex] 
 \hline\hline
 Ours            & 92.8 & 89.1 & 79.7 & 70.0 &78.9 &73.2 &66.7 &82.9 &78.9 &31 \\ 
 \hline
 \cite{deepcut1} & 92.4 & 88.9 & 79.1 &67.9 &78.7 &72.4 &65.4 &82.1 &78.1 &270 \\
\end{tabular}
\caption{We display mean average precision (mAP) of our  approach versus the baseline of \cite{deepcut1} for the various human parts as well as whole body.}
 \label{pose-estimation}
\end{center}
\end{table*}
%
%
%
%

%
\subsection{Cell Instance Segmentation}
\label{ssec:cell}
\begin{figure*}[]
\begin{center}
\includegraphics[angle=0, width=0.9\textwidth]{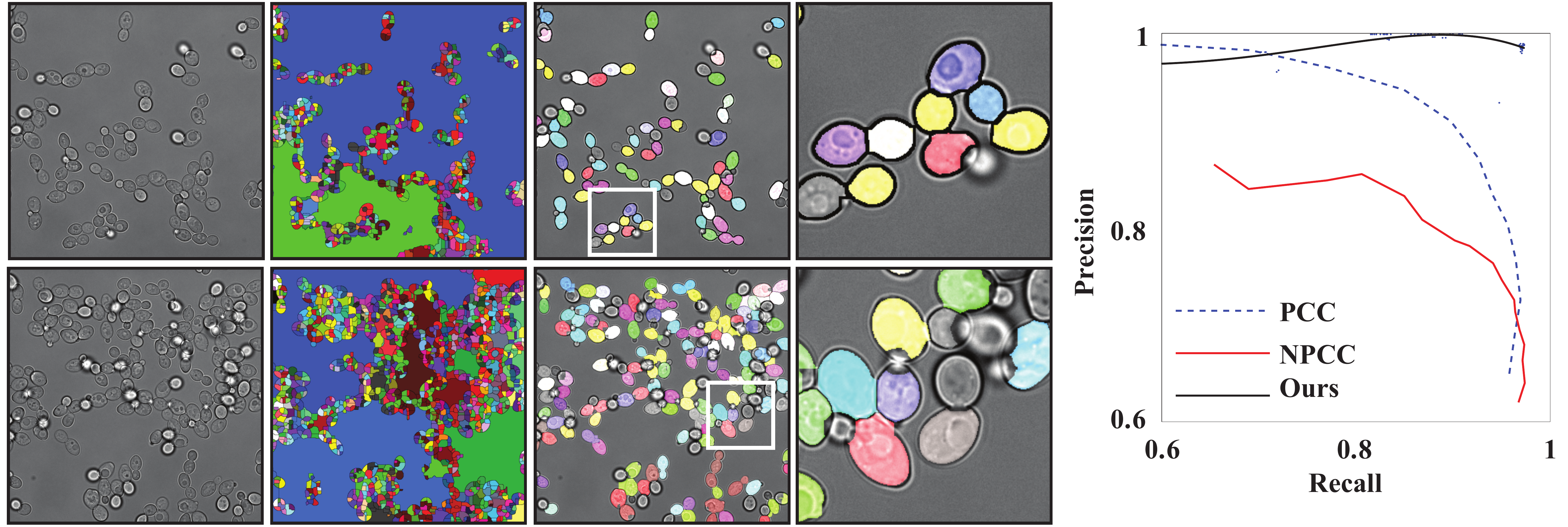}
  \caption[]{(\emph{left}) Example cell segmentation results. Columns are (\emph{left} to \emph{right}): original image, super-pixels, color map of segmentation, and enlarged views of the inset (\emph{white frame}). (\emph{right}) Precision-Recall plot of the cell detection, compared with those using the planar correlation clustering (PCC) and the non-planar correlation clustering (NPCC) techniques in~\cite{Zhang14b}.  We generate the precision-recall plot by using various offsets to $\Gamma$ to generate segmentations with more or less cells.  }
  \label{fig:seg_examples_cells}
  \end{center}
\end{figure*}
The technique described in this paper is applicable to images that have crowded cell regions acquired from different modalities and cell types, while individual cells are mainly discernible by boundary cues. Our approach starts by generating an over-segmentation, or a set of super-pixels, then it finds a clustering of the super-pixels into an arbitrary number of segmented regions, i.e. cells. 

We use the bright field microscopy of yeast cells data set from~\cite{Zhang14a}. 
Challenges of this data set include: densely packed and touching cells, out-of-focus artifacts, variations on shape and size, changing boundaries even on the same cell, as well as other structures showing similar boundaries. We train a logistic regression classifier to determine $\phi$ which we then offset by a hand tuned constant fixed for the data set.  For each super-pixel $d$, the proportion of it in the background defined $\theta_d$ which we then offset by a hand tuned constant fixed for the data set.   

We compare the performance of our algorithm with current state-of-the-art method ~\cite{Zhang14b}. 
Here ~\cite{Zhang14b} uses the algorithms planar correlation clustering (PCC) and non-planar correlation clustering (NPCC) . The precision-recall plot is shown in Fig.~\ref{fig:seg_examples_cells}, where our detection rate is higher than both PCC and NPCC. The Jaccard Index values are: 86.3$\pm$13.3, 86.4$\pm$12.0 and 90.3$\pm$8.5 for PCC, NPCC and our technique, respectively.
On a data set of 1635 images, 85.07 \% had a gap of zero and hence were solved exactly.  With regards to the normalized gaps (99.3400,99.2800   99.0800,98.8200,95.72)\% had normalized gap under (0.16,0.1,0.01,0.001,.0001) respectively. We tighten the bound using odd set inequalities of size three which are discussed in the supplement.  We plot a histogram of inference time in Fig \ref{histplotpose}.   
%
%
\section{Conclusion}
\label{concsectiondone}
We introduce new formulations of multi-person pose segmentation and cell instance segmentation.  Given these formulations we introduce novel inference algorithms designed to attack the problems.  Our algorithms use column generation and our models are structured so that generating columns can be done efficiently.  We compare our results to the state of the art algorithms on multi-person pose segmentation and cell instance segmentation.  We demonstrate that our algorithms rapidly produce more accurate results than the baseline. 
\section*{Acknowledgment}
\label{ack}
This work is partly supported by the Spanish Ministry of Economy and Competitiveness under the Maria de Maeztu Units of Excellence Programme (MDM-2015-0502).
\clearpage
{\small
\bibliographystyle{ieee}
\bibliography{egbib,bib_pose,cell_segmentation}
}
\clearpage
\appendix
\section{Overview of Supplement}
In this supplement we consider the problems of tightening the LP relaxation beyond that in the main paper, anytime lower bounds and bounds on $\lambda$.  In Section \ref{betterLP} we study the problem of tightening the LP relaxation for instance segmentation then apply that knowledge to pose segmentation.  

In Section \ref{explower} we derive anytime lower bounds on instance segmentation then apply that knowledge to pose segmentation.    In Section \ref{bounPoseBase} we study a set of bounds on the Lagrange multipliers $\lambda^1,\lambda^4,\lambda^5$ in pose segmentation.
\section{Tightening the Bound}
\label{betterLP}
Our original LP relaxation only contains constraints for collections of poses/instances
that share a common key-point.  This generally results in a tight relaxation in our domain though not always.  We correct this issue using odd set inequalities  \cite{heismann2014generalization}.  Specifically we use odd set inequalities of size three.  In practice the introduction of these inequalities results in a tight relaxation.  

This section is organized as follows.  In Section \ref{dispProb} we show a simple case where the relaxation in the main paper  $\min_{\substack{\gamma \geq 0 \\ G\gamma \leq 1}}\Gamma^t \gamma$ is loose.  Next in Section \ref{3fix} we show how to correct this using odd set inequalities of order three which we refer to as triples.  Then in Section \ref{pdform} we formulate optimization over the tighter relaxation.  Next in Section \ref{pdalg} we adapt our column generation algorithm to the new tighter bound. In Section \ref{rowgen} we show how to add odd set inequalities to optimization.  Finally in Section \ref{colgentri} we show how to generate columns.  In Section \ref{tripPose}-\ref{colTripPosemake} we study the corresponding tighter bound with regards to pose segmentation.   
\subsection{Fractional Solutions from Mutually Exclusive Triples}
\label{dispProb}
We now consider a case where the LP relaxation $\min_{\substack{\gamma \geq 0 \\ G\gamma \leq 1}}\Gamma^t \gamma$ provides a loose lower bound. We use the identical case from \cite{yarkoNips2016}.  Consider  a set $\mathcal{Q}=\{q_1,q_2,q_3,q_4\}$ over $\mathcal{D}=\{ d_1,d_2,d_3\}$ where the first three members of $\mathcal{Q}$ each contain two of three members of $\mathcal{D}$, $\{d_1,d_2\},\{d_1,d_3\},\{d_2,d_3\}$, and
the fourth  contains all three $\{d_1,d_2,d_3\}$.  Suppose the corresponding costs
are given by $\Gamma_{q_1}=\Gamma_{q_2}=\Gamma_{q_3}=-4$ and $\Gamma_{q_4}=-5$.
The optimal integer solution sets $\gamma_{q_4}=1$, and has a cost of
$-5$.  However a lower cost fractional solution sets
$\gamma_{q_1}=\gamma_{q_2}=\gamma_{q_3}=0.5$;
$\gamma_{q_4}=0$ which has cost $-6$.  Hence the LP relaxation is loose
in this case.  Even worse, rounding the fractional solution results in a
sub-optimal solution. 
\subsection{Tightening the Bound over Triples of Key-Points}
\label{3fix}
A tighter bound can be motivated by the following observation.  For any set of
three unique members of $\mathcal{D}$ denoted $d_1,d_2,d_3$ the number of selected members of $\mathcal{Q}$ that include two or more members of $d_1,d_2,d_3$ can be no larger than one. We write this observation below
\begin{align}
\label{lpcorrect3}
\sum_{q \in \mathcal{Q}}[G_{d_1q}+G_{d_2q}+G_{d_3q} \geq 2]\gamma_q\leq 1 
\end{align}
We now apply our tighter bound to instance segmentation.  We denote the set of sets of three unique members of $\mathcal{D}$ as $\mathcal{C}$ and index it with $c$.   For short hand we refer to $\mathcal{C}$ as the set of triples.  Following the notation of \cite{yarkoNips2016} we describe $\mathcal{C}$ with constraint matrix $C \in \{0,1\}^{|\mathcal{C}|\times|\mathcal{Q}|}$ which we index by $c,q$.  Here $C_{cq}=1$ if and only if column $q$ contains two or more members of set $c$.  We define $C$ formally below.  
\begin{align}
C_{cq}=[\sum_{d \in c}G_{dq}\geq 2] \quad \forall c \in \mathcal{C}, q \in \mathcal{Q}
\end{align}
Given $C$ we write  a tighter bound on $\gamma$ than in the main paper below.
\begin{align}
\min_{\substack{\gamma \in \{0,1\}^{|\mathcal{Q}|} \\ G\gamma \leq 1}}\Gamma^t\gamma
\geq
\min_{\substack{\gamma \geq 0 \\ G\gamma \leq 1 \\ C\gamma \leq 1}}\Gamma^t\gamma 
\geq \min_{\substack{\gamma \geq 0 \\ G\gamma \leq 1 }}\Gamma^t\gamma
\end{align}
\subsection{Primal and Dual Formulation}
\label{pdform}
The primal and dual LP relaxations of instance segmentation with constraints corresponding to  Eq \ref{lpcorrect3} are written below.  The dual is expressed using Lagrange multipliers $\lambda \in \mathbb{R}_{0+}^{|\mathcal{D}|}$ and  $\kappa \in \mathbb{R}_{0+}^{|\mathcal{C}|}$.
\begin{align}
\label{betterRelax}
\min_{\substack{\gamma \geq 0 \\ G\gamma \leq 1 \\ C\gamma \leq 1 }}\Gamma^t\gamma 
= \max_{\substack{\lambda \geq 0 \\ \kappa \geq 0 \\ \Gamma+G^t\lambda+C^t\kappa \geq 0}}1^t\lambda +1^t\kappa 
\end{align}
\subsection{Algorithm}
\label{pdalg}
We now attack optimization of Eq \ref{betterRelax}.  Column generation on its own is not sufficient for this task as the number of rows in $C$ (members of $\mathcal{C}$) scales cubically in the cardinality of $\mathcal{D}$ and hence we can not use the entire matrix $C$ during optimization.   However empirically we observe that very few rows of $C$ are needed to ensure an integral solution.  This motivates the use of the cutting plane method to build a sufficient subset of $\mathcal{C}$ jointly while applying column generation.  The joint procedure is called column/row generation.  

We denote the nascent subsets of $\mathcal{Q},\mathcal{C}$
as $\hat{\mathcal{Q}}, \hat{\mathcal{C}}$ respectively.
In Alg \ref{cyccell} write column/row generation optimization given subroutines $\mbox{COLUMN}(\lambda,\kappa)$,$\mbox{ROW}(\gamma)$ that identify a group of violated constraints in
primal and dual including the most violated in each.  We  define the subroutines in Sections \ref{rowgen},\ref{colgentri}.
 \begin{algorithm}[H]
 \caption{Column/Row Generation}
\begin{algorithmic} 
\State $\hat{\mathcal{Q}} \leftarrow \{ \},\quad \hat{\mathcal{C}} \leftarrow \{ \}$
\Repeat
\State $[\lambda,\kappa]\leftarrow \max_{\substack{\lambda \geq 0\\ 
\kappa\geq 0 \\
\Gamma_{\hat{\mathcal{Q}}}+G_{(:,\hat{\mathcal{Q}})}^t\lambda
+ C_{(\mathcal{\hat{C}},\mathcal{\hat{Q}})}^t\kappa \geq 0}}
    -1^t\lambda-1^t\kappa $
\State Recover $\gamma$ from $\lambda,\kappa$
\State $\dot{\mathcal{Q}} \leftarrow \mbox{COLUMN}(\lambda,\kappa) $
\State $ \dot{\mathcal{C}} \leftarrow \mbox{ROW}(\gamma)$
%
\State  $\hat{\mathcal{Q}}\leftarrow [\hat{\mathcal{Q}},\dot{\mathcal{Q}}]$
\State   $\hat{\mathcal{C}}\leftarrow [\hat{\mathcal{C}},\dot{\mathcal{C}}]$
 \Until{ $\dot{\mathcal{Q}}=[]$  and $\dot{\mathcal{C}}=[]$ }
\end{algorithmic}
\label{cyccell}
  \end{algorithm}
\subsection{Row Generation}
\label{rowgen}
Finding the most violated row consists of the following optimization.  
\begin{align}
\max_{c \in \mathcal{C}}\sum_{q \in \mathcal{Q}} C_{cq} \gamma_q
\end{align}
Enumerating $\mathcal{C}$ is unnecessary and we generate its rows as needed by considering
only $c=\{ d_{c_1} d_{c_2} d_{c_3} \}$ such that for each of pair $d_{c_i},d_{c_j}$  there exists an index $q$ such that $\gamma_q>0$ and $G_{d_iq}=G_{d_jq}=1$.  Generating rows is done only when no (significantly) violated  columns exist.
%
%
Triples are only added to $\hat{\mathcal{C}}$ if the corresponding constraint is violated.  
\subsection{Generating Columns}
\label{colgentri}
We apply the column generation procedure of Section \ref{colbb} to generate columns.  The  ILP is modified in Eq \ref{getcolcell} by introducing $\kappa$ terms into the objective. 
\begin{align}
\min_{\substack{x \in \{ 0,1\}^{|\mathcal{Q}|} \\ x_{d_*}=1}} &\sum_{d \in \mathcal{D}}(\theta_d+\lambda_dx_d+\theta_d)+\sum_{d_1,d_2 \in \mathcal{D}}\phi_{d_1d_2}x_{d_1}x_{d_2} +\sum_{c \in \mathcal{C}}\kappa_c([2 \leq \sum_{d \in c}x_d])\\
\nonumber & x_d =0 \quad \forall d\in \mathcal{D} \quad \mbox { s.t. }L_{d,d_*}>m_R\\
\nonumber &\sum_{d\in \mathcal{D}}x_d \leq m_V
\end{align}
\subsection{Tighter Bound for Multi-Person Pose Segmentation}
\label{tripPose}
The technique of Eq \ref{betterRelax} can be applied in the context of our work on multi-person pose segmentation.  The corresponding constraints that are to be enforced are the following.  No more than one global pose can include more than two members of a given set of three key-points.  No more than one local pose can include more than two members of a given set of three key-points (either as local or global).  We formalize this below.

We use $C^{\mathcal{L}}\in \{ 0,1\}^{|\mathcal{C}| \times |\mathcal{L}|}$ to define the adjacency matrix between triples and local poses .  Similarly we use $C^{\mathcal{G}}\in \{ 0,1\}^{|\mathcal{C}| \times |\mathcal{G}|}$ to define the adjacency matrix between triples and global poses.  Here $C^{\mathcal{L}}_{cq}=1$ if and only if  local pose $q$ contains two or more members of set $c$.  Similarly we set $C^{\mathcal{G}}_{cq}=1$ if and only if global pose $q$ contains two or more members of set $c$.
We define $C^{\mathcal{L}},C^{\mathcal{G}}$ formally below.  
\begin{align}
&C^{\mathcal{G}}_{cq}=[(\sum_{d \in c}G_{dq})\geq 2] &\quad \forall c \in \mathcal{C}, q \in \mathcal{G}\\
\nonumber &C^{\mathcal{L}}_{cq}=[(\sum_{d \in c}L_{dq}+M_{dq})\geq 2] &\quad \forall c \in \mathcal{C}, q \in \mathcal{L}
\end{align}
\subsection{Dual Form}
We now write the corresponding primal LP for multi-person pose segmentation with triples added. 
\begin{align}
\label{primalTriRelax}
\mbox{Eq } \ref{primdualilp} \geq \min_{\substack{\gamma\geq 0 \\ \psi \geq 0 \\ G\gamma+L\psi \leq 1 \\ L\psi+M\psi\leq 1 \\-G\gamma+M\psi\leq 0 \\C^{\mathcal{G}}\gamma \leq 1 \\ C^{\mathcal{L}}\psi \leq 1}}\Gamma^t\gamma+\Psi^t \psi 
\geq \mbox{Eq } \ref{dualFormPose}
\end{align}
We now take the dual of the central term in inequality in Eq \ref{primalTriRelax}.  This induces two additional sets of Lagrange multipliers $\lambda^4,\lambda^5 \in \mathbb{R}_{0+}^{\mathcal{C}}$.  We now write the dual below.
\begin{align}
 \min_{\substack{\gamma\geq 0 \\ \psi \geq 0 \\ G\gamma+L\psi \leq 1 \\ L\psi+M\psi\leq 1 \\-G\gamma+M\psi\leq 0 \\C^{\mathcal{G}}\gamma \leq 1 \\ C^{\mathcal{L}}\psi \leq 1}}\Gamma^t\gamma+\Psi^t \psi 
 %
%
=\max_{\substack{\lambda \geq 0 \\  \Gamma+G^t(\lambda^1-\lambda^3)+C^{\mathcal{G}t}\lambda^4 \geq 0 \\ \Psi+L^t\lambda^1+(M^t+L^t)\lambda^2+M^t\lambda^3 +C^{\mathcal{L}t}\lambda^5 \geq 0}} -1^t\lambda^1-1^t\lambda^2-1^t\lambda^3-1^t\lambda^4-1^t\lambda^5
\label{dualForwTrip}
\end{align}
\subsection{Algorithm}
Optimization proceeds for multi-person pose segmentation exactly as for instance segmentation in Section \ref{pdalg}.  We need only describe the procedure for generating violated rows and columns.  We write the optimization algorithm below.  
 \begin{algorithm}[H]
 \caption{Column/Row Generation}
\begin{algorithmic} 
\State $\hat{\mathcal{G}} \leftarrow \{ \}$
\State $\hat{\mathcal{L}} \leftarrow \{ \}$
\State $ \hat{\mathcal{C}} \leftarrow \{ \}$
\Repeat
\State $[\lambda]\leftarrow $ Maximize dual in Eq \ref{dualForwTrip} over column and rows sets $\hat{\mathcal{G}},\hat{\mathcal{L}},\hat{\mathcal{C}}$
\State Recover $\gamma$ from $\lambda$
\State $\dot{\mathcal{G}},\dot{\mathcal{L}} \leftarrow \mbox{COLUMN}(\lambda) $
\State $ \dot{\mathcal{C}} \leftarrow \mbox{ROW}(\gamma,\psi)$
%
\State  $\hat{\mathcal{G}}\leftarrow [\hat{\mathcal{G}},\dot{\mathcal{G}}]$
\State  $\hat{\mathcal{L}}\leftarrow [\hat{\mathcal{L}},\dot{\mathcal{L}}]$
\State   $\hat{\mathcal{C}}\leftarrow [\hat{\mathcal{C}},\dot{\mathcal{C}}]$
 \Until{ $\dot{\mathcal{G}}=[]$ and $\dot{\mathcal{L}}=[]$  and $\dot{\mathcal{C}}=[]$ }
\end{algorithmic}
\label{cyccPOSE}
\end{algorithm}
%
%
\subsection{Generating rows}
Generating rows corresponding to local poses is done separately for each part. We write the correpsonding optimization for identifying the most violated constraint corresponding to a local pose over a given part $r$ as follows.  
\begin{align}
\max_{\substack{c \in \mathcal{C}\\ R_d=r \; \; \forall d\in c }}\sum_{q \in \mathcal{L}} C^{\mathcal{L}}_{cq} \psi_q
\end{align}
Finding violated triples for global poses is assisted by the knowledge that one need only consider triples over three unique part types as no global pose includes two or more key-points of a given part.  Hence only such triples need be considered for global pose.  The corresponding optimization is below. For any given $c$ let the key-points associated with it be described as follows $c=\{ d_{c_1} d_{c_2} d_{c_3} \}$.
\begin{align}
\max_{\substack{c \in \mathcal{C}\\ R_{d_{c_1}}\neq R_{d_{c_2}}\\ R_{d_{c_1}}\neq R_{d_{c_3}}\\  R_{d_{c_2}}\neq R_{d_{c_3}}}}\sum_{q \in \mathcal{G}} C^{\mathcal{G}}_{cq} \gamma_q
\end{align}
Triples are only added to $\hat{\mathcal{C}}$ if the corresponding constraint is violated.  
\subsection{Generating Columns}
\label{colTripPosemake}
Generating columns is considered separately for global and local poses.  The corresponding equations are unmodified from Section \ref{violposes} except for the introduction of terms over triples. 
%
We write the ILP for generating the most violated local pose  given the global key-point below. 
\begin{align}
\label{makeColTripL}
\min_{\substack{x \in \{ 0,1\}^{|\mathcal{D}|} \\x_{d_*}=1}}(-\lambda^1_{d_*}+\lambda^3_{d_*}-\theta_{d_*}) +\sum_{d \in \mathcal{D}}(\theta_d+\lambda^1_{d}+\lambda^2_d)x_d
 +\sum_{d_1,d_2 \in \mathcal{D}}x_{d_1}x_{d_2}\phi_{d_1d_2}+\sum_{c \in \mathcal{C}}\lambda^5_{cq}[ \sum_{d \in c}x_{d}  \geq 2]
\end{align}
%
%
For each $d_*$ such that $R_{d_*} \in \hat{\mathcal{R}}$ we compute the most violated constraint corresponding to a global pose including $d_*$.  We write this as an ILP below.
\begin{align}
\label{makeColTripG}
\min_{ \substack{z \in \{ 0,1\}^{|\mathcal{D}|} \\ z_{d_*}=1\\  \sum_{d \in \mathcal{D}}[R_d=r]z_d \leq 1 \; \; \forall r \in \mathcal{R}}}\sum_{ d\in \mathcal{D}} (\theta_{d}+\lambda^1_d-\lambda^3_d)z_{d}
 +\sum_{d_1d_2 \in \mathcal{D}} \phi_{d_1d_2}z_{d_1}z_{d_2}+\sum_{c \in \mathcal{C}}\lambda^4_{cq}[ \sum_{d \in c}z_{d}  \geq 2]
\end{align}
The introduction of triples breaks problem structure that is a precondition to attack optimization via dynamic programming.  However we found that directly solving with an ILP solver is not problematic computationally for our problems.
\section{Lower Bounds for Cell Instance Segmentation}
\label{explower}
We now consider the computation of an anytime lower bound on the optimal cell instance segmentation.  We first write the ILP for cell instance segmentation and then introduce Lagrange multipliers.  
 \begin{align}
\min_{\substack{\gamma \in \{0,1\}^{|\mathcal{Q}|} \\ G\gamma \leq 1 \\ C\gamma \leq 1 }}\Gamma^t\gamma  
= \min_{\substack{\gamma \in \{0,1\}^{|\mathcal{Q}|} \\G\gamma \leq 1 }}\max_{\substack{\lambda \geq 0 \\ \kappa \geq 0}}\Gamma^t\gamma+(-\lambda^t1+\lambda^tG\gamma)+(-\kappa^t1+\kappa^tC\gamma)
\label{LBcell0}
 \end{align}
 We now relax the constraint in Eq \ref{LBcell0} that the dual variables are optimal producing the following lower bound.
 \begin{align}
\nonumber \mbox{Eq }\ref{LBcell0} &\geq  \min_{\substack{\gamma \in \{0,1\}^{|\mathcal{Q}|}\\ G\gamma \leq 1 }}\Gamma^t\gamma+(-\lambda^t1+\lambda^tG\gamma)-(\kappa^t1+\kappa^tC\gamma) \\
 &=-\kappa^t1-\lambda^t1+\min_{\substack{\gamma \in \{0,1\}^{|\mathcal{Q}|}\\ G\gamma \leq 1 }}(\Gamma+G^t\lambda+C^t\kappa)^t\gamma
 \label{LBcell1}
 \end{align}
Recall that every cell is associated with at least one centroid.  We denote the set of centroids associated with a given cell $q$ as $\mathcal{N}_{q}$.  Given any fixed $\gamma\in \{0,1\}^{|\mathcal{Q}|}$ such that $G\gamma \leq1$ observe the following.  
\begin{align}
(\Gamma+G^t\lambda+C^t\kappa)^t\gamma \geq \sum_{d \in \mathcal{D}}\min[0,\min_{\substack{q\in \mathcal{Q} \\ d \in \mathcal{N}_q}}\gamma_{q}(\Gamma_q+G_{:,q}^t\lambda+C_{:,q}^t\kappa)]
\label{boundtool}
\end{align}
%
We now use  Eq \ref{boundtool} to produce following lower bound on Eq \ref{LBcell1}. 
\begin{align}
\mbox{Eq } \ref{LBcell1}\geq -\kappa^t1-\lambda^t1+\min_{\substack{\gamma \in \{0,1\}^{|\mathcal{Q}|}\\ G\gamma \leq 1 }}\sum_{d \in \mathcal{D}}\min[0,\min_{\substack{q\in \mathcal{Q} \\ d \in \mathcal{N}_q}}\gamma_{q}(\Gamma_q+G_{:,q}^t\lambda+C_{:,q}^t\kappa)]
\label{LBcell2}
\end{align}
We now relax the constraint in Eq \ref{LBcell2} that $G\gamma \leq 1$ producing the following lower bound.  
\begin{align}
\mbox{Eq } \ref{LBcell2}\geq & -\kappa^t1-\lambda^t1+\min_{\substack{\gamma \in \{0,1\}^{|\mathcal{Q}|} }}\sum_{d \in \mathcal{D}}\min[0,\min_{\substack{q\in \mathcal{Q} \\ d \in \mathcal{N}_q}}\gamma_{q}(\Gamma_q+G_{:,q}^t\lambda+C_{:,q}^t\kappa)]\\
\nonumber =&-\kappa^t1-\lambda^t1+\sum_{d \in \mathcal{D}}\min[0,\min_{\substack{q\in \mathcal{Q} \\ d \in \mathcal{N}_q}}(\Gamma_q+G_{:,q}^t\lambda+C_{:,q}^t\kappa)]
\end{align}
Observe that the term $\min_{\substack{q\in \mathcal{Q} \\ d \in \mathcal{N}_q}}(\Gamma_q+G_{:,q}^t\lambda+C_{:,q}^t\kappa)$ is identical to the optimization computed at every stage of column/row generation.  To get the bounds in Section \ref{lowerplacmain} one simply ignores the $\kappa$ terms (setting them to zero) as triples are not considered in the main paper.  
\subsection{Discussion of Lower Bounds: Pose}
\label{lbapp}
 We now consider computing an anytime lower bound on the multi-person pose segmentation.  We first write the ILP  for multi-person pose segmentation.  
\begin{align}
\label{step1poe}
\min_{\substack{\gamma \in \{0,1\}^{\mathcal{G}} \\ \psi \in \{0,1\}^{\mathcal{L}} \\ G\gamma+L\psi \leq 1 \\ L\psi+M\psi\leq 1 \\-G\gamma+M\psi\leq 0 \\C^{\mathcal{G}}\gamma \leq 1 \\ C^{\mathcal{L}}\psi \leq 1}}\Gamma^t\gamma+\Psi^t \psi 
\end{align}
We now augment this with the following two redundant constraints.  No key-point can be the global key-point in more than one selected local pose.  Furthermore no key-point corresponding to a global part can be present in more than one global pose. We use $G_{\hat{\mathcal{R}}}$ to describe the rows of matrix G corresponding to key-points associated with a global part. 
\begin{align}
\mbox{Eq }\ref{step1poe}= \min_{\substack{\gamma \in \{0,1\}^{\mathcal{G}} \\ \psi \in \{0,1\}^{\mathcal{L}}  \\ G\gamma+L\psi \leq 1 \\ L\psi+M\psi\leq 1 \\-G\gamma+M\psi\leq 0 \\C^{\mathcal{G}}\gamma \leq 1 \\ C^{\mathcal{L}}\psi \leq 1 \\ M\psi \leq 1 \\ G_{\hat{\mathcal{R}}} \gamma \leq 1}}\Gamma^t\gamma+\Psi^t \psi 
\end{align}
We now replace constraints with Lagrange multipliers except for the new constraints.  
\begin{align}
\mbox{Eq } \ref{step1poe}=&
\min_{\substack{\gamma \in \{0,1\}^{\mathcal{G}} \\ \psi \in \{0,1\}^{\mathcal{L}} \\M\psi \leq 1 \\  G_{\hat{\mathcal{R}}} \gamma \leq 1}}
\max_{\lambda \geq 0} \lambda^{1t}(G\gamma-L\psi-1)+\lambda^{2t}(L\psi+M\psi-1) \\
\nonumber &+ \lambda^{3t}(-G\gamma+M\psi)+\lambda^{4t}(C^{\mathcal{G}}\gamma-1)+\lambda^{5t}(C^{\mathcal{L}}\psi-1)+\Gamma^t\gamma+\Psi^t\psi
\end{align}
We now relax optimality in $\lambda$ producing a lower bound.  
\begin{align}
\label{my_inq_pose}
\mbox{Eq } \ref{step1poe} &\geq  -1^t\lambda^1-1^t\lambda^2-1^t\lambda^3-1^t\lambda^4-1^t\lambda^5 \\
\nonumber &+\min_{\substack{\gamma \in \{0,1\}^{\mathcal{G}} \\ \psi \in \{0,1\}^{\mathcal{L}} \\M\psi \leq 1 \\ G_{\hat{\mathcal{R}}}\gamma \leq 1}}(\Gamma+G^t\lambda^1-G^t\lambda^3+C^{\mathcal{G}t}\lambda^4)^t\gamma+(L^t\lambda^1+L^t\lambda^2+M^t\lambda^2+M^t\lambda^3+C^{\mathcal{L}t}\lambda^5)\psi\\
\nonumber &=-1^t\lambda^1-1^t\lambda^2-1^t\lambda^3-1^t\lambda^4-1^t\lambda^5 \\
\nonumber &+\min_{\substack{\gamma \in \{0,1\}^{\mathcal{G}} \\ G_{\hat{\mathcal{R}}}\gamma \leq 1}}(\Gamma+G^t\lambda^1-G^t\lambda^3+C^{\mathcal{G}t}\lambda^4)^t\gamma \\
\nonumber &+\min_{\substack{\psi \in \{0,1\}^{\mathcal{L}}\\ M\psi \leq 1}}
(L^t\lambda^1+L^t\lambda^2+M^t\lambda^2+M^t\lambda^3+C^{\mathcal{L}t}\lambda^5)^t\psi
\end{align}
Observe that the constraint $M\psi\leq 1$ only requires that no key-point is the global part in more than one local pose. We alter Eq \ref{my_inq_pose} is account for this below.  
\begin{align}
\label{my_inq_pose2}
\mbox{Eq }\ref{my_inq_pose}& = -1^t\lambda^1-1^t\lambda^2-1^t\lambda^3-1^t\lambda^4-1^t\lambda^5 \\
\nonumber &+\min_{\substack{\gamma \in \{0,1\}^{\mathcal{G}} \\ G_{\hat{\mathcal{R}}}\gamma \leq 1}}(\Gamma+G^t\lambda^1-G^t\lambda^3+C^{\mathcal{G}t}\lambda^4)^t\gamma \\
\nonumber &+\sum_{d_* \in \mathcal{D}} \min[0,\min_{\substack{q \in \mathcal{L}\\ M_{d_*q}=1}}
(L^t_{:,q}\lambda^1+L^t_{:,q}\lambda^2+M^t_{:,q}\lambda^2+M^t_{:,q}\lambda^3+C^{\mathcal{L}t}_{:,q}\lambda^5)]
\end{align}
Recall that every global pose is associated with at least one global part.  Given any fixed $\gamma\in \{0,1\}^{|\mathcal{G}|}$ such that $G\gamma \leq1$ observe the following.  
\begin{align}
\label{myfixtransfomr}
(\Gamma+G^t\lambda^1-G^t\lambda^3+C^{\mathcal{G}}\lambda^4)^t\gamma \geq 
\sum_{\substack{d_* \in \mathcal{D} \\ R_d\in \hat{\mathcal{R}}}}\min [ 0,\min_{\substack{q \in \mathcal{G} \\  G_{d_*q}=1 }} (\Gamma_q+G^t_{:,q}(\lambda^1-\lambda^3)+C^{\mathcal{G}}_{:,q}\lambda^4)\gamma_q]
\end{align}
We now apply Eq \ref{myfixtransfomr} to produce a relaxation of Eq \ref{my_inq_pose2}.    
\begin{align}
\label{posebound2}
\mbox{Eq }\ref{my_inq_pose2} & \geq  -1^t\lambda^1-1^t\lambda^2-1^t\lambda^3-1^t\lambda^4-1^t\lambda^5 \\
\nonumber &+\min_{\substack{\gamma \in \{0,1\}^{\mathcal{G}} \\ G_{\hat{\mathcal{R}}}\gamma \leq 1}} \sum_{\substack{d_* \in \mathcal{D} \\ R_{d_*}\in \hat{\mathcal{R}}}}\min [ 0,\min_{\substack{q \in \mathcal{G} \\  G_{d_*q}=1 }} (\Gamma_q+G^t_{:,q}(\lambda^1-\lambda^3)+C^{\mathcal{G}}_{:,q}\lambda^4)\gamma_q] \\
\nonumber &+\sum_{d_* \in \mathcal{D}} \min[0,\min_{\substack{q \in \mathcal{L}\\ M_{d_*q}=1}}
(L^t_{:,q}\lambda^1+L^t_{:,q}\lambda^2+M^t_{:,q}\lambda^2+M^t_{:,q}\lambda^3+C^{\mathcal{L}t}_{:,q}\lambda^5)]
\end{align}
We relax the constraint that $G_{\hat{\mathcal{R}}}\gamma \leq 1$ to produce the following bound.   
%
\begin{align}
\label{posebound3}
\mbox{Eq }\ref{posebound2} &\geq  -1^t\lambda^1-1^t\lambda^2-1^t\lambda^3-1^t\lambda^4-1^t\lambda^5 \\
\nonumber &+\sum_{\substack{d \in \mathcal{D} \\ R_d\in \hat{\mathcal{R}}}}\min [ 0,\min_{\substack{q \in \mathcal{G} \\  G_{dq}=1 }} \Gamma_q+G^t_{:,q}(\lambda^1-\lambda^3)+C^{\mathcal{G}}_{:,q}\lambda^4] \\
\nonumber &+\sum_{d_* \in \mathcal{D}} \min[0,\min_{\substack{q \in \mathcal{L}\\ M_{d_*q}=1}}
(L^t_{:,q}\lambda^1+L^t_{:,q}\lambda^2+M^t_{:,q}\lambda^2+M^t_{:,q}\lambda^3+C^{\mathcal{L}t}_{:,q}\lambda^5)]
\end{align}
Now observe that the two minimizations in Eq \ref{posebound3} are the identical minimizations used when generating columns under triples in Eq \ref{makeColTripG} and Eq \ref{makeColTripL} which we solve during each step of column/row generation.  Thus Eq \ref{posebound3} describes an anytime tractable lower bound.
\section{Bounding the Lagrange Multipliers For Multi-Person Pose Segmentation}
\label{bounPoseBase}
We now study the problem of bounding Lagrange multipliers so as to induce speed ups in dual optimization.
Inspired by the work of \cite{PlanarCC,HPlanarCC}, we observe that prior to convergence of column generation that the optimal solution may not lie in the span of the set of columns produced thus far.   It is useful to allow some values of $G\gamma+L\psi$  to exceed one.  The following situation motivates this:  Consider that we have two very low cost global poses and that they overlap at a single key-point $d_1$ corresponding to a non-global part.  Furthermore assume that the strength of the pairwise connections between $d_1$ and its neighbors are very weak.  It preferable to use both global poses and simply forget that $d_1$ was included in one of the poses.  However the hard constraint that $G\gamma+L\psi\leq 1$ prevents the use of both in the solution. 

To permit the violation of $G\gamma+L\psi\leq 1$ we introduce a slack vector $\omega^1 \in \mathbb{R}_{0+}^{|\mathcal{D}|}$ that tracks the presence of any ``over-included" key-points  and prevents them from making a  negative contribution to the objective. Similarly we introduce slack vectors $\omega^4 \in  \mathbb{R}_+^{|\mathcal{C}^{\mathcal{G}}|},$ and $\omega^5  \in  \mathbb{R}_+^{|\mathcal{C}^{\mathcal{L}}|}$ to allow for the constraints $C^{\mathcal{G}}\gamma \leq 1$ and $C^{\mathcal{L}}\psi \leq 1$ to be violated.  We  associate vectors $\omega^1$,$\omega^4$,$\omega^5$ with cost vectors $\Omega^1 \in \mathbb{R}_+^{|\mathcal{D}|},\Omega^4 \in  \mathbb{R}_+^{|\mathcal{C}^{\mathcal{G}}|},$ and $\Omega^5  \in  \mathbb{R}_+^{|\mathcal{C}^{\mathcal{L}}|}$ respectively.  

We replace the constraint $G\gamma+L\psi\leq 1$ with $G\gamma+L\psi -\omega^1 \leq 1$ and associate an additional cost $\Omega^{1t}\omega^1$ to the objective.  
Similarly we replace $C^{\mathcal{G}}\gamma \leq 1$ and $C^{\mathcal{L}}\psi \leq 1$ with $C^{\mathcal{G}}\gamma -\omega^4 \leq 1$, and $C^{\mathcal{L}}\gamma -\omega^5 \leq 1$ respectively.  We add the following cost to the objective $\Omega^{4t}\omega^4+ \Omega^{5t}\omega^5$.  We define $\Omega^1,\Omega^4,\Omega^5$ such that it is the case at termination of column generation that the optimal solution sets $\omega^1,\omega^4,\omega^5$ to zero.  
%
%
We write the primal and dual LPs of optimization augmented with $\omega,\Omega$ terms below.  
\begin{align}
\label{primalVerTripBOund}
\mbox{Eq }\ref{dualForwTrip}=
\min_{\substack{\gamma \geq 0 \\ \psi \geq 0\\ \omega \geq 0}}&\Gamma^t \gamma +\Psi^t \psi+ \Omega^{1t}\omega^1+\Omega^{4t}\omega^4+\Omega^{5t}\omega^5 \\
\nonumber & G^t\gamma+L^t\psi -\omega^1\leq 1 \\
\nonumber &M^t \psi+L^t\psi \leq 1\\
\nonumber &-G^t\gamma +M^t\psi\leq 0\\
\nonumber & C^{\mathcal{G}}\gamma -\omega^4\leq 1\\
\nonumber & C^{\mathcal{L}}\psi -\omega^5 \leq 1\\
=
\max_{\lambda \geq 0 } &-1^t\lambda^1-1^t\lambda^2-1^t\lambda^3-1^t\lambda^4-1^t\lambda^5 \\
\nonumber &\Gamma+G^t(\lambda^1-\lambda^3)+C^{\mathcal{G}t}\lambda^4 \geq 0 \\
\nonumber &\Psi+L^t\lambda^1+(M^t+L^t)\lambda^2+M^t\lambda^3 +C^{\mathcal{L}t}\lambda^5 \geq 0 \\
\nonumber &\Omega^1\geq \lambda^1\\
\nonumber &\Omega^4\geq \lambda^4\\
\nonumber &\Omega^5\geq \lambda^5
\end{align}
The result of the introduction of these slack terms in the primal is bounds on the dual variables $\lambda^1,\lambda^4,\lambda^5$.
%
The remainder of this section is devoted to determining the values of $\Omega^1,\Omega^4,\Omega^5$.  One could trivially set all $\Omega$ terms to be infinite.  However we found empirically that tighter bounds speed optimization.  This phenomena is observed in \cite{PlanarCC,HPlanarCC}.  

We determine $\Omega$ by creating an algorithm that projects a feasible solution to Eq \ref{primalVerTripBOund} with non-zero $\omega$ terms to a feasible solution with zero valued $\omega$ terms.  This algorithm provides for the computation of upper bounds on decrease in the cost achieved by the projection.  Given this upper bound we simply set $\Omega$ terms so that the changes in the objective induced by the projection are guaranteed to be negative.   It should be observed that this algorithm is for analysis only and is not in our code.  It is only used to establish the values $\Omega^1,\Omega^4,\Omega^5$.  
%

We develop this algorithm in Section \ref{drop1} and apply it to determine $\Omega^1$.  In Section \ref{drop2} and Section \ref{drop3} we apply it to determine $\Omega^5$ and $\Omega^4$ respectively. 
\subsection{Computing $\Omega^1$}
\label{drop1}
In this subsection we consider the construction of a bound on $\Omega^1$.  This bound is defined to be infinite for all $d$ associated with a global part and non infinite otherwise.    
We now consider a procedure that iterates through $d_*$ such that $R_{d_*}\notin \hat{\mathcal{R}}$ and $\omega^1_{d_*}>0$.  
After each iteration  our procedure updates terms in $\gamma,\psi$ so as to achieve the following goals  for a given $d_* \in \mathcal{D}$.
\begin{itemize}
\item Decrease the objective in Eq \ref{primalVerTripBOund}
\item Remain feasible for Eq \ref{primalVerTripBOund}.
\item Decrease or leave constant all $\omega$ terms
\item Set $\omega^1_{d_*}$ to zero
\end{itemize} 
For any selection of $d_* \in \mathcal{D}$ we  consider a mapping where a given pose $q$ is mapped to a corresponding pose $\bar{q}$.  The domain of our mapping is the set of global poses where $G_{d_*q}=1$ and local poses where $L_{d_*q}=1$. For a given global pose $q$ such that $G_{d_*q}=1$ the corresponding pose $\bar{q}$ is identical to $q$ except that $G_{d_*\bar{q}}=0$.  For a given local pose $q$ such that $L_{d_*q}=1$ the corresponding pose $\bar{q}$ is identical to q except that $L_{d_*\bar{q}}=0$.  We define this mapping below.  
%
%
%
%
\begin{align}
&G_{d\bar{q}}=G_{dq}[d\neq d_*]&\quad \forall d \in \mathcal{D}, q\in \mathcal{G}\quad  G_{d_*q}=1\\
\nonumber &M_{d\bar{q}}=M_{dq} &\quad \forall d \in \mathcal{D}, q\in \mathcal{L}\quad L_{d_*q}=1\\
\nonumber &L_{d\bar{q}}=L_{dq}[d\neq d_*]&\quad \forall d \in \mathcal{D}, q\in \mathcal{L}\quad L_{d_*q}=1
\end{align}
%
%
We now consider our update to $\psi,\gamma$.   
After the updates are applied we set $\omega^1_{d_*}=0$. The updates that achieve this are written below.  
\begin{align}
\label{updateEquDelta}
\gamma^{\mbox{NEW}}_q= &\gamma_q-\gamma_q\frac{\omega^1_{d_*}}{1+\omega^1_{d_*}} &\quad  \forall q \in \mathcal{G}, G_{d_*q}=1\\
\nonumber \psi^{\mbox{NEW}}_q=& \psi_q-\psi_q\frac{\omega^1_{d_*}}{1+\omega^1_{d_*}} & \quad  \forall q \in  \mathcal{L};\; (M_{d_*q}+L_{d_*q}=1) \\
\nonumber  \gamma^{\mbox{NEW}}_{\bar{q}}=& \gamma_{\bar{q}}+\gamma_q\frac{\omega^1_{d_*}}{1+\omega^1_{d_*}}& \quad  \forall q \in \mathcal{G};G_{d_*q}=1\\
\nonumber \psi^{\mbox{NEW}}_{\bar{q}}=&\psi_{\bar{q}}+ \psi_q\frac{\omega^1_{d_*}}{1+\omega^1_{d_*}}[\Psi_{\bar{q}}<0] & \quad  \forall q \in \mathcal{L};\quad L_{d_*q}=1
\end{align}
These updates describe why we only consider $d_* \in \mathcal{R}_d$.  Recall that a global pose must include a key-point corresponding to a global part.  If the updates in Eq \ref{updateEquDelta} were applied on $d_*$ where $d_*\in \hat{ \mathcal{R}}$ then global poses that have a no key-point corresponding to a global part could be introduced into the solution.  Hence we set $\Omega^1_{d}=\infty$ for all $d$ such that $R_d\in \hat{\mathcal{R}}$.  This ensures that the corresponding terms in $\omega^1$ are set to zero by optimization.  

After the updates in Eq \ref{updateEquDelta} are applied for any given $d_* \in \mathcal{D}$ it may be the case that $\omega$ terms can be decreased leaving $\gamma$ and $\psi$ fixed.  These updates decrease objective if they modify any terms and are written below.  
\begin{align}
\label{updateOmegaEZ}
\omega^1_d\leftarrow & \max[0,G_{d,:}\gamma^{\mbox{NEW}}+L_{d,:}\psi^{\mbox{NEW}}-1] &\quad \forall d \in \mathcal{D}\\
\nonumber \omega^4_c \leftarrow &\max[0,C^{\mathcal{G}}_c\gamma^{\mbox{NEW}}-1] &\quad \forall c \in \mathcal{C}^{\mathcal{G}}\\
\nonumber \omega^5_c \leftarrow &\max[0,C^{\mathcal{L}}_c\psi^{\mbox{NEW}}-1]& \quad  \forall c \in \mathcal{C}^{\mathcal{L}}
\end{align}
\subsubsection{Establishing the value of  $\Omega^1_{d_*}$}
\label{observeOmega}
We denote the  change in the objective achieved by the updates in Eq \ref{updateEquDelta}  as $\Delta$ which we define below. %
\begin{align}
\nonumber \Delta=-\omega^1_{d_*}\Omega^1_{d_*}+\sum_{q\in \mathcal{G}}\frac{\omega^1_{d_*}}{1+\omega^1_{d_*}}\gamma_qG_{d_*q}(\Gamma_{\bar{q}}-\Gamma_q)+\sum_{q\in \mathcal{L}}\frac{\omega^1_{d_*}}{1+\omega^1_{d_*}}\psi_q(L_{d_*q}+M_{d_*q})(\Psi_{\bar{q}}L_{d*q}[\Psi_{\bar{q}}<0]-\Psi_q)\\
=\frac{\omega^1_{d_*}}{1+\omega^1_{d_*}}(-(1+\omega^1_{d_*})\Omega^1_{d_*}+\sum_{\substack{q\in \mathcal{G}\\G_{d_*q}=1}}\gamma_q(\Gamma_{\bar{q}}-\Gamma_q)+\sum_{\substack{q\in \mathcal{L}\\ L_{d_*q}=1}}\psi_q(\Psi_{\bar{q}}[\Psi_{\bar{q}}<0]-\Psi_q)+\sum_{\substack{q\in \mathcal{L}\\ M_{d_*q}=1}}\psi_q(0-\Psi_q))
\label{DeltaEq}
\end{align}
From Eq \ref{DeltaEq} we know that in order for $\Delta$ to be negative the following must hold for any solution $\gamma,\psi$.
\begin{align}
\label{mustsatis}
(1+\omega^1_{d_*})\Omega^1_{d_*}>\sum_{\substack{q\in \mathcal{G}\\G_{d_*q}=1}}\gamma_q(\Gamma_{\bar{q}}-\Gamma_q)+\sum_{\substack{q\in \mathcal{L}\\ L_{d_*q}=1}}\psi_q(\Psi_{\bar{q}}[\Psi_{\bar{q}}<0]-\Psi_q)+\sum_{\substack{q\in \mathcal{L}\\ M_{d_*q}=1}}\psi_q(0-\Psi_q))
\end{align}
Thus Eq \ref{mustsatis} defines the conditions for a suitable  $\Omega^1_{d_*}$. To identify a suitable value $\Omega^1_{d_*}$ we produce an upper bound on Eq \ref{DeltaEq} and set $\Omega^1_{d_*}$ accordingly.  The resultant value on  $\Omega^1_{d_*}$ term is strictly greater than the term needed to satisfy Eq \ref{mustsatis}.  

We now proceed to upper bound Eq \ref{DeltaEq}.  We achieve this using terms $\alpha_1,\alpha_2,\alpha_3 \in \mathbb{R}_{0+}$ which are defined below.
Observe that $\max[0,\max_{\substack{q\in \mathcal{G}\\G_{d_*q}=1}}(\Gamma_{\bar{q}}-\Gamma_q)] =\max[0,\max_{\substack{q\in \mathcal{G}\\G_{d_*q}=1}}-\theta_{d_*}+\sum_{d_1\neq d}\phi_{d_*d_1}G_{d_1q}] $ which we upper bound with $\alpha_1$ as follows. 
\begin{align}
%
\alpha_1=\max[0,-\theta_{d_*}-\sum_{\substack{r \in \mathcal{R}\\ r \neq R_{d_*}}}\min[0,  \min_{\substack{d_1 \in \mathcal{D}\\R_{d_1}=r}} \phi_{d_*d_1}]]
\end{align}
We define $\alpha_2,\alpha_3$ below.  
\begin{align}
 & \alpha_2= \max[0,\max_{\substack{q\in \mathcal{L}\\L_{d_*q}=1}}(\Psi_{\bar{q}}[\Psi_{\bar{q}}<0]-\Psi_q)] \\
\nonumber & \alpha_3= \max[0,\max_{\substack{q\in \mathcal{L}\\M_{d_*q}=1}}((0-\Psi_q))]
\end{align}
Observe that $\alpha_1$ can be computed by checking each pairwise term including key-point $d_*$.  We compute $\alpha_2,\alpha_3$ via exhaustive search.  The computation of  $\alpha_2,\alpha_3$ is tractable because we do not  have more than fifteen key-points per part in any problems in our data set and hence we only have up to $2^{15}$ possible $q$ to consider.  
We now upper bound   Eq \ref{DeltaEq} as follows.  
\begin{align}
\label{deltaStep2}
\mbox{Eq } \ref{DeltaEq} \leq  \frac{\omega^1_{d_*}}{1+\omega^1_{d_*}}(-(1+\omega^1_{d_*})\Omega^1_{d_*}+\alpha_1\sum_{\substack{q\in \mathcal{G}\\G_{d_*q}=1}}\gamma_q+\alpha_2\sum_{\substack{q\in \mathcal{L}\\ L_{d_*q}=1}}\psi_q+\alpha_3\sum_{\substack{q\in \mathcal{L}\\ M_{d_*q}=1}}\gamma_q)
\end{align}
Given that $-G_{d,:}\gamma+M_{d,:}\psi \leq 0$ and that  $\alpha_3\geq 0$ we conclude the following. 
\begin{align}
\label{boundTrick}
\alpha_3\sum_{\substack{q\in \mathcal{L}\\ M_{d_*q}=1}}\gamma_q\leq \alpha_3\sum_{\substack{q\in \mathcal{G}\\G_{d_*q}=1}}\gamma_q
\end{align}
%
We now use Eq \ref{boundTrick} to upper bound Eq \ref{deltaStep2} . 
\begin{align}
\label{deltaStep3}
\mbox{Eq }\ref{deltaStep2} \leq \frac{\omega^1_{d_*}}{1+\omega^1_{d_*}}(-(1+\omega^1_{d_*})\Omega^1_{d_*}+(\alpha_1+\alpha_3)\sum_{\substack{q\in \mathcal{G}\\G_{d_*q}=1}}  \gamma_q+\alpha_2 \sum_{\substack{q\in \mathcal{L}\\ L_{d_*q}=1}}\psi_q)
\end{align}
We now upper bound Eq \ref{deltaStep3} by  replacing $\alpha_1+\alpha_3$ and $\alpha_2$ with $\max[\alpha_1+\alpha_3,\alpha_2]$.  
\begin{align}
\label{deltaStep4}
\mbox{Eq }\ref{deltaStep3} \leq 
 \frac{\omega^1_{d_*}}{1+\omega^1_{d_*}}(-(1+\omega^1_{d_*})\Omega^1_{d_*}+ \max[\alpha_1+\alpha_3,\alpha_2](\sum_{\substack{q\in \mathcal{L}\\ L_{d_*q}=1}}\psi_q+\sum_{\substack{q\in \mathcal{G}\\G_{d_*q}=1}} \gamma_q))
\end{align}
Recall that $G_{d_*,:}\gamma+L_{d_*,:}\psi \leq 1+\omega^1_{d_*}$ and that $\max[\alpha_1+\alpha_3,\alpha_2]$ is non-negative. We therefor establish the following.    %
\begin{align}
 \max[\alpha_1+\alpha_3,\alpha_2](\sum_{\substack{q\in \mathcal{L}\\ L_{d_*q}=1}}\psi_q+\sum_{\substack{q\in \mathcal{G}\\G_{d_*q}=1}} \gamma_q)) \leq  \max[\alpha_1+\alpha_3,\alpha_2] (1+\omega^1_{d_*})
\end{align}
We now upper bound Eq \ref{deltaStep4} by replacing $G_{d_*,:}\gamma+L_{d_*,:}\psi$ with $1+\omega^1_{d_*}$.%
%
\begin{align}
\mbox{Eq }\ref{deltaStep4}=& \frac{\omega^1_{d_*}}{1+\omega^1_{d_*}}(-(1+\omega^1_{d_*})\Omega^1_{d_*}+ \max[\alpha_1+\alpha_3,\alpha_2](1+\omega^1_{d_*}))\\
\nonumber =& \omega^1_{d_*}(-\Omega^1_{d_*}+ \max[\alpha_1+\alpha_3,\alpha_2])
\end{align}
In order to ensure that $\Delta <0$ it is sufficient to ensure that $\Omega^1_{d_*}> \max[\alpha_1+\alpha_3,\alpha_2])$.  Let $\epsilon$ be a tiny positive number.  We now define $\Omega^1_{d_*}$.  
\begin{align}
\Omega^1_d=\epsilon+\max[\alpha_1+\alpha_3,\alpha_2]
\end{align}
%
%
\subsection{Computing $\Omega^5$}
\label{drop2}
%
%
In this subsection we consider the construction of a bound on $\Omega^5$. 
We now consider a procedure that iterates through $c$ such that $\omega^5_c>0$.  
After each iteration  our procedure updates terms in $\gamma,\psi$ so as to achieve the following goals  for a given $c \in \mathcal{C}$.
\begin{itemize}
\item Decrease the objective in Eq \ref{primalVerTripBOund}
\item Remain feasible for Eq \ref{primalVerTripBOund}.
\item Decrease or leave constant all $\omega$ terms
\item Set $\omega^5_{c}$ to zero
\end{itemize} 
We now consider a mapping of local poses to other local poses.  The domain of our mapping is the set of local poses where members $q$ satisfy $C^{\mathcal{L}}_{cq}=1$. For a given local pose $q$ such that $C^{\mathcal{L}}_{cq}=1$ the corresponding pose $\bar{q}$ is identical to $q$ except that $L_{d\bar{q}}=0$ for all $d \in c$.  We define this mapping below.  %
%
\begin{align}
&M_{d\bar{q}}=M_{dq} \quad & \forall d\in \mathcal{D},q \in \mathcal{L}, \quad C^{\mathcal{L}}_{cq}=1\\
\nonumber &L_{d\bar{q}}=L_{dq}[d \notin c] \quad &\forall d\in \mathcal{D},q \in \mathcal{L} \quad C^{\mathcal{L}}_{cq}=1
\end{align}
We now consider our update to $\psi,\gamma$.   
After the updates are applied we set $\omega^5_{c}=0$. The updates that achieve this are written below.  
%
%
\begin{align}
\label{updateEquDelta4}
&\psi^{\mbox{NEW}}_q= \psi_q-\psi_q\frac{\omega^5_{c}}{1+\omega^5_{c}} \quad & \forall q \in  \mathcal{L};\; C^{\mathcal{L}}_{cq}=1\\
\nonumber &\psi^{\mbox{NEW}}_{\bar{q}}=\psi_{\bar{q}}+ \psi_q\frac{\omega^5_{c}}{1+\omega^5_{c}}[\Psi_{\bar{q}}<0] \quad& \forall q \in \mathcal{L}; C^{\mathcal{L}}_{cq}=1
\end{align}
Our iterative procedure may cause some constraints defining  $\omega^1,\omega^4,\omega^5$ to become loose for active $\omega$ terms.  Thus we decrease $\omega$ terms according to Eq \ref{updateOmegaEZ} without increasing the objective.  
\subsubsection{Establishing the value of  $\Omega^5_c$}
\label{observeOmega4}
We denote the  change in the objective achieved by the updates in Eq \ref{updateEquDelta4}  as $\Delta$.   We define $\Delta$ below in terms of the $\omega,\psi $ terms before the update in Eq \ref{updateEquDelta4} is applied for $c$.
%
%
\begin{align}
\label{DeltaEqLOC}
\Delta=-\omega_{c}\Omega_{c}+\sum_{\substack{q\in \mathcal{L}\\ C^{\mathcal{L}}_{cq}=1}}\frac{\omega_{c}}{1+\omega_{c}}\psi_q(\Psi_{\bar{q}}[\Psi_{\bar{q}}<0]-\Psi_q)
\end{align}
We now proceed to upper bound Eq \ref{DeltaEqLOC}.  We achieve this using term $\alpha$ defined as follows.
\begin{align}
\alpha=\max[0,\max_{\substack{q \in \mathcal{L}\\ C^{\mathcal{L}}_{cq}=1}}(\Psi_{\bar{q}}[\Psi_{\bar{q}}<0]-\Psi_q)]
\end{align} 
We compute $\alpha$ via exhaustive search.  This is tractable in our data set since each part is associated with less than fifteen key-points in general and usually much less.  
Using $\alpha $ we bound Eq \ref{DeltaEqLOC} as follows.  
\begin{align}
\label{DeltaEqLOC2}
\mbox{Eq }\ref{DeltaEqLOC}\leq \omega_{c}\Omega_{c}+\sum_{\substack{q\in \mathcal{L}\\ C^{\mathcal{L}}_{cq}=1}}\frac{\omega_{c}}{1+\omega_{c}}\psi_q\alpha
\end{align}
Recall that  $\alpha$ is non-negative and that $C^{\mathcal{L}}_{c:}\psi \leq \omega^5+1$.  Thus the following holds.  
\begin{align}
\label{localTrick}
\alpha C^{\mathcal{L}}_{c:}\psi \leq \alpha(\omega^5_c+1)
\end{align}
Using Eq \ref{localTrick} we bound Eq \ref{DeltaEqLOC2} as follows.  
\begin{align}
\label{DeltaEqLOC3}
\mbox{Eq }\ref{DeltaEqLOC2}\leq 
-\omega_{c}\Omega_{c}+\alpha \omega_c
\end{align}
Thus in order to ensure that our updates in Eq \ref{updateEquDelta4} decrease the objective we define $\Omega^5_c$ as follows using tiny positive real number $\epsilon$.  
\begin{align}
\Omega^5_{c}=\epsilon+\alpha
\end{align}
%
\subsection{Computing $\Omega^4$}
\label{drop3}
We now consider the construction of a bound on $\Omega^4$.  This bound is defined to be infinite for all $c$ associated more than one global part and non infinite otherwise.    
We now consider a procedure that iterates through $c$ such that $c$ contains no more than one key-point corresponding to a global part and $\omega^4_c>0$.   

Let $c$ be defined by key-points $\{d_1,d_2,d_3\}$.  If one key-point corresponding to a global part is present in $c$ then that key-point is associated with $d_1$ otherwise the assignment of key-points in $c$ to indexes $d_1,d_2,d_3$ is done arbitrarily.

After each iteration  our procedure updates terms in $\gamma,\psi$ so as to achieve the following goals  for a given $d_* \in \mathcal{D}$.
\begin{itemize}
\item Decrease the objective in Eq \ref{primalVerTripBOund}
\item Remain feasible for Eq \ref{primalVerTripBOund}.
\item Decrease or leave constant all $\omega$ terms
\item Set $\omega^4_{c}$ to zero
\end{itemize} 
For any selection of $c \in \mathcal{C}$ we  consider a mapping where a given pose $q$ is mapped to a corresponding $\bar{q}$. The domain of our mapping is the set of global poses where $C^{\mathcal{G}}_{cq}=1$. For a given global pose $q$ such that $C^{\mathcal{G}}_{cq}=1$ the corresponding pose $\bar{q}$ is identical to $q$ except that $G_{d_2\bar{q}}=G_{d_3\bar{q}}=0$.  We define this mapping below.  
\begin{align}
G_{d\bar{q}}=G_{dq}[d\neq d_2][d\neq d_3]] \quad \forall d \in \mathcal{D}, C^{\mathcal{G}}_{cq}=1
\end{align}
%
We now consider our update to $\psi,\gamma$. After the updates are applied we set $\omega^4_{c}=0$. The updates that achieve this are written below.  
\begin{align}
\label{updateEquDelta5}
\gamma^{\mbox{NEW}}_q=& \gamma_q-\gamma_q\frac{\omega^4_{c}}{1+\omega^4_c} \quad  \forall q \in \mathcal{G}, G_{d_*q}=1\\
\nonumber \psi^{\mbox{NEW}}_q=& \psi_q-\psi_q\frac{\omega^4_c}{1+\omega^4_c} \quad \forall q \in  \mathcal{L}, [M_{d_2q}+M_{d_3q}=1]\\
\nonumber  \gamma^{\mbox{NEW}}_{\bar{q}}=& \gamma_{\bar{q}}+\gamma_q\frac{\omega^4_c}{1+\omega^4_c}\quad \forall q \in \mathcal{G};G_{d_*q}=1
\end{align}
These updates describe why we only consider $c \in \mathcal{C}$ where no more than one part is global.  Recall that a global pose must include at least one key-point corresponding to a global part.  If the updates in Eq \ref{updateEquDelta4} were applied on $c$ where where two or more key-points correspond to global parts then global poses that have a no key-point corresponding to a global part could be introduced into the solution.  Hence we set $\Omega^4_{c}=\infty$ for all $c$ where two or more of the key-points correspond to global parts.  This ensures that the corresponding terms in $\omega^4$ are set to zero by optimization.  
\subsubsection{Establishing the value of  $\Omega^4_{c}$}
We denote the  change in the objective achieved by the updates in Eq \ref{updateEquDelta5}  as $\Delta$ which we define below. %
\begin{align}
\label{DeltaEq5}
\Delta=-\omega^4_c \Omega^4_c+\sum_{q \in \mathcal{G}} \gamma_qC^{\mathcal{G}}_{cq}\frac{\omega^4_c}{1+\omega^4_c}(\Gamma_{\bar{q}}-\Gamma_{q})
+\sum_{\substack{q \in \mathcal{Q}\\M_{d_2q}=1}} \psi_q\frac{\omega^4_c}{1+\omega^4_c}(-\Psi_q)
+\sum_{\substack{q \in \mathcal{Q}\\M_{d_3q}=1}} \psi_q\frac{\omega^4_c}{1+\omega^4_c}(-\Psi_q)
\end{align}

We now proceed to upper bound Eq \ref{DeltaEq5}.  We achieve this using terms $\alpha_1,\alpha_2,\alpha_3 \in \mathbb{R}_{0+}$ defined as follows. We define $\alpha_1$ as an upper bound on $\max_{\substack{q \in \mathcal{Q}\\ C^{\mathcal{G}}_{cq}=1}}(\Psi_{\bar{q}}-\Psi_{q})$ as follows.
\begin{align}
\alpha_1=\max [0, -\theta_{d_2}- \theta_{d_3}-\phi_{d_1d_2}-\phi_{d_1d_3}-\phi_{d_2d_3}\\
\nonumber - \sum_{r \in \mathcal{R}-R_{d_1}-R_{d_2}-R{d_3}}\min[0,\min_{\substack{d\in \mathcal{D}\\R_d=r}}\phi_{d_2d} ] \\
\nonumber -\sum_{r \in \mathcal{R}-R_{d_1}-R_{d_2}-R{d_3}}\min[0,\min_{\substack{d\in \mathcal{D}\\R_d=r}}\phi_{d_3d} ]]
\end{align}
%
%
%
We define $\alpha_2,\alpha_3$ below.
\begin{align}
\nonumber \alpha_2=\max(0,\max_{\substack{q \in \mathcal{L}\\M_{d_2q}=1}}-\Psi_q)\\
\nonumber \alpha_3=\max(0,\max_{\substack{q \in \mathcal{L}\\M_{d_3q}=1}}-\Psi_q)
\end{align}
Using $\alpha_1,\alpha_2,\alpha_3$ we produce the following bound on Eq \ref{DeltaEq5}.  
\begin{align}
\label{globBound1}
\Delta \leq -\omega^4_c \Omega^4_c+\alpha_1\sum_{q \in \mathcal{G}} \gamma_q\frac{\omega^4_c}{1+\omega^4_c}+\alpha_2\sum_{\substack{q \in \mathcal{L}\\M_{d_2q}=1}}\psi_q\frac{\omega^4_c}{1+\omega^4_c}
+\alpha_3\sum_{\substack{q \in \mathcal{L}\\M_{d_3q}=1}}\psi_q\frac{\omega^4_c}{1+\omega^4_c}
\end{align}
Recall that for all  $d\in \mathcal{D}$ that $\sum_{q\in \mathcal{L}}L_{dq}+ M_{dq}\leq 1$ and that $\alpha_2,\alpha_3$ are non-negative.  Therefor the following hold.  
\begin{align}
\label{my5trick}
\alpha_2 \sum_{\substack{q \in \mathcal{L}\\M_{d_2q}=1}}  \psi_q \leq \alpha_2\\
\nonumber \alpha_3\sum_{\substack{q \in \mathcal{L}\\M_{d_3q}=1}}\psi_q \leq \alpha_3
\end{align}
Using Eq \ref{my5trick} we bound  Eq \ref{globBound1} as follows.  
\begin{align}
\label{globBound2}
\mbox{Eq }\ref{globBound1}\leq -\omega^4_c \Omega^4_c+\sum_{\substack{q \in \mathcal{G}\\C^{\mathcal{G}}_{cq}=1}} \gamma_q\frac{\omega^4_c}{1+\omega^4_c}\alpha_1+\alpha_2\frac{\omega^4_c}{1+\omega^4_c}+\alpha_3\frac{\omega^4_c}{1+\omega^4_c}
\end{align}
Recall that  $C^{\mathcal{G}}_{c:}\gamma \leq 1+\omega^4_c$ and that $\alpha_1$ is non-negative.  Thus we bound Eq \ref{globBound2} as follows.  
\begin{align}
\label{globBound3}
\mbox{Eq }\ref{globBound2}&\leq-\omega^4_c \Omega^4_c+\alpha_1 \omega^4_c+(\alpha_2+\alpha_3)\frac{\omega^4_c}{1+\omega^4_c}\\
\nonumber &\leq  -\omega^4_c \Omega^4_c+\alpha_1 \omega^4_c+(\alpha_2+\alpha_3)\omega^4_c\\
\nonumber &=\omega^4_c(-\Omega^4_c+\alpha_1+\alpha_2+\alpha_3)
\end{align}
In order to ensure that $\Delta <0$ it is sufficient to ensure that $\Omega^4_c> \alpha_1+\alpha_2+\alpha_3$.  Thus we set $\Omega^4_c$ as follows. 
\begin{align}
\Omega^4_c=\epsilon+\alpha_1+\alpha_2+\alpha_3
\end{align}
Recall that when all three key-points in $c$ do not correspond to global parts then our allocation of members of $c$ to $d_1,d_2,d_3$ is arbitrary.  However our bound also indicates that the choice of which member corresponds to $d_1$ does alter the value of the bound.  Thus the tightest bound can be determined by trying all choices for $d_1$ and computing the corresponding bound for each. 
%
%
%
%
%
\end{document}